# The Illinois Social Attitudes Aggregate Corpus (ISAAC): An Open Tool and Reproducible Pipeline for Analyzing Social Group Discourse at Scale

Babak Hemmatian[1], Sarah Hadjarab[2], Jessica Chen[3], and Benedek Kurdi[3, 4]

[1] AI Innovation Institute, Stony Brook University

[2] Department of Political Science, University of Chicago

[3] Department of Psychology, University of Illinois Urbana–Champaign

[4] Social & Behavioral Science Institute, University of Illinois Urbana–Champaign

## Author Note

Babak Hemmatian https://orcid.org/0000-0001-6138-5782

Sarah Hadjarab https://orcid.org/0009-0003-7382-5433

Jessica Chen https://orcid.org/0009-0000-3578-195X

Benedek Kurdi https://orcid.org/0000-0001-5000-0584

We have no conflicts of interest to disclose. This work was supported by startup funds provided by the University of Illinois Urbana–Champaign to B.K.

Correspondence concerning this article should be addressed to Babak Hemmatian, 100 Nicolls Rd, Stony Brook, NY 11794, United States. Email: babak.hemmatian@stonybrook.edu.

**Abstract**

We introduce the Illinois Social Attitudes Aggregate Corpus (ISAAC), an open, modular, and accessible corpus of 527 million+ English-language Reddit posts selected for relevance to six key social group distinctions based on race, sexuality, age, ability, body weight, and skin tone, covering the 17-year period from 2007 to 2023. A multi-step, human-audited filtering pipeline was used to keep irrelevant content in the curated dataset below 10%, both overall and for each social group distinction. Each post was then algorithmically annotated with the user's estimated home region, along with a suite of validated off-the-shelf and custom semantic labels including moralization, sentiment, emotion, and linguistic generalization. We confirm the validity of the resulting corpus through convergent evidence linking ISAAC to macro-level societal trends, such as online search behavior, temporal spikes during major societal events (both nationally and regionally), and long-term shifts in public attitudes. By offering a unified, public infrastructure, ISAAC eliminates research fragmentation and enables seamless replication while supporting diverse empirical workflows at scale. Specifically, ISAAC allows investigators to perform cross-category comparisons, conduct high-precision tracking of long-term temporal shifts in social group discourse, and map spatial variation onto localized public opinion and policy outcomes. ISAAC's fully public, modular pipeline facilitates easy extension of the corpus to new platforms, languages, and social categories. To accommodate various research needs, ISAAC is accessible both without coding through a point-and-click website and labeler web-apps, and programmatically via an SQL playground, a Python package, and HuggingFace.

# The Illinois Social Attitudes Aggregate Corpus (ISAAC): An Open Tool and Reproducible Pipeline for Analyzing Social Group Discourse at Scale

Social group attitudes have been a central topic of inquiry across the social sciences, including social psychology (Allport, 1954; Wood, 2000), sociology (Blumer, 1958; Link & Phelan, 2001), political science (Kinder & Kam, 2009; Kinder & Sanders, 1996), communication (Gerbner et al., 1980; Schiappa et al., 2005), and economics (Becker, 1957; Manski, 2000). Beyond its core meaning referring to evaluations of social categories such as age, race, and sexual orientation along a positive–negative continuum (Eagly & Chaiken, 1993), here we use the term expansively to encompass any form of social group-related cognition and affect to match our multi-dimensional empirical approach. As such, our definition also covers social group stereotypes (Fiske et al., 2002), intergroup emotions (Cottrell & Neuberg, 2005), and social group identity (Tajfel & Turner, 1986), among others.

The methods that social sciences bring to the study of social group attitudes are varied and include self-report surveys (e.g., Sears & Henry, 2003), long-running trend series (e.g., Brooks & Bolzendahl, 2004), laboratory experiments (e.g., Tajfel et al., 1971), indirect measures (e.g., Greenwald et al., 1998), vignette studies (e.g., Rossi et al., 1974), conjoint analysis (e.g., Caruso et al., 2009), field and audit studies (e.g., Bertrand & Mullainathan, 2004), and qualitative analysis of media content (e.g., Dixon & Williams, 2015). Together, these diverse disciplinary perspectives and methodologies have produced a rich body of knowledge about the origins of social group attitudes, their basic operation, and consequences for individual behavior, institutions, and policy.

For all their differences, traditional approaches share the key commonality of observing attitudes under conditions elicited by the researcher: a questionnaire administered, a stimulus presented, a scenario described, or a conversation recorded. However, with the advent of computational social science (Lazer et al., 2009; Lazer et al., 2020) comes the possibility of studying the same attitudes in more organic, naturalistic, and externally valid settings. Large public archives of online discourse now record billions of unprompted statements that people make about themselves, each other, and the social categories that organize their lives. The computational advances that have fueled progress in cognitive modeling (Binz et al., 2025; Griffiths, 2015) and natural language processing (NLP; Devlin et al., 2019; Vaswani et al., 2017) have made at-scale analyses of such spontaneously produced discourse tractable.

Together, these computational advances open large-scale, unstructured social discourse to rigorous quantitative analysis, allowing investigators to gain empirical insight into long-standing questions. How are social group attitudes expressed in spontaneous communication? How do they shift in the weeks, months, and years around significant societal events? How and when does discourse change from describing particular individuals to making claims about entire social categories? How does the quantity and linguistic structure of social group discourse relate to other, established attitudinal indicators, such as measures of explicit and implicit evaluation?

However, answering these questions requires a level of methodological consistency that current practices cannot support. As empirical fields of study mature, their cumulative progress depends on transitioning away from siloed discoveries toward shared methodological standards, infrastructure, and benchmarks. Without unified baselines, a field's growth becomes structurally constrained; individual research teams operate in isolation, making it difficult if not impossible to distinguish genuine scientific breakthroughs from noise associated with idiosyncratic aspects of an individual study. This issue is especially acute for computational fields, where the complexity, opacity, and flexibility of preprocessing pipelines routinely involve dozens of researcher decisions at every stage.

Presently, the computational study of social group attitudes suffers from this vulnerability: each new study typically scrapes its own corpus, selects its own filters, applies its own labelers, and validates against its own benchmarks. The resulting methodological heterogeneity ultimately thwarts systematic replication. For example, if two separate studies of online discourse about race report different patterns of results over the same time period, it is unclear whether the discrepancy reflects a substantive change in real-world discourse, platform-specific noise, or the operationalization choices made in each pipeline (Olteanu et al., 2019; Ruths & Pfeffer, 2014).

Different disciplines have faced similar fragmentation in their maturation and successfully resolved it by converging on shared, validated infrastructure. For example, in computer vision, the transition to unified benchmarks like ImageNet (Russakovsky et al., 2015) catalyzed modern machine learning by establishing a standard metric for progress. Psycholinguistics, developmental psychology, and network science also stabilized their empirical baselines through common datasets and resources like the British National Corpus (Burnard, 2007), the CHILDES database (MacWhinney, 2000), and the SNAP datasets (Leskovec & Krevl, 2014).

A parallel transition to standardized infrastructure and an open, shared pipeline is now overdue for the computational study of social group-related discourse. In the absence of such a resource, four recurring weaknesses will continue to limit the scope and validity of the inferences — weaknesses that the Illinois Social Attitudes Aggregate Corpus (ISAAC) was designed to address.

First, researchers routinely isolate relevant content using keyword filters that capture unrelated senses of a word — such as "black" in a chess match or "disabled" in a software update — while rarely reporting false positive rates (Olteanu et al., 2019; Ruths & Pfeffer, 2014; Tufekci, 2014). Irrelevant content introduces statistical noise, inflating standard errors and thereby compromising the reliability, validity, and statistical power of downstream analyses. These issues underscore a clear need for more sophisticated data filtering pipelines.

Second, semantic labels like sentiment and emotion are typically generated using singular (and often one-off) models, ignoring how training data and model behaviors vary (Hutto & Gilbert, 2014; Qi et al., 2020; see also Chan et al., 2021; van Atteveldt et al., 2021). The baseline tendencies of models are so strong and divergent that they can produce diametrically opposed results when applied to the same data. For example, of the three sentiment tools we evaluate below, the proportion of identical posts labeled negative ranges from 20% to 64%. Furthermore, if a model's training data associates a specific social group with hostile language, it reproduces that bias as a measurement artifact. As such, an apparent difference between groups may be due to the labeler rather than the discourse itself. For some semantic dimensions like linguistic generalization, the problem is even more acute: because there are no widely accessible models, each researcher is left to devise their own methodology. Relying on the same, broad set of labelers with redundancies and minimal theoretical commitments across studies mitigates these forms of error.

Third, validation against established empirical metrics, such as measures of implicit and explicit evaluation, remains rare (but see Caliskan et al., 2017; Garg et al., 2018). Any quantitative indicator must be validated by testing its relationship with independent benchmarks (Campbell & Fiske, 1959), and computational paradigms are no exception. Without convergent validation, the foundational construct validity of text-based attitudinal measures remains unknown. An unvalidated corpus marker that fluctuates over time is subject to ambiguity: apparent changes may stem from shifts in users, evolving platform norms, or internal drift within the classification tool, not a meaningful change in the construct of interest.

Fourth, the rare studies that do link text measures to external attitude indicators (such as Caliskan et al., 2017; Garg et al., 2018) rely on highly curated, historical, or artificial materials, such as published books or crowd-written probe sentences. For instance, the historical share of scientific text in Google Books grew markedly across the twentieth century (Pechenick et al., 2015), while probe sentences are explicitly elicited rather than volunteered. Curated datasets reflect elite or formal discourse, which can diverge sharply from the everyday, informal language that most social science work seeks to target (Broockman & Skovron, 2018; Zaller, 1992). Consequently, a validation pattern established in curated, formal archives may fail to hold in the organic, spontaneous environments where public attitudes are actively expressed.

Several existing resources have addressed individual aspects of these vulnerabilities. Some corpora, including naturalistic Reddit and Twitter dumps such as the Arctic Shift archive (Heitmann, 2026) and the Twitter Decahose (Fafalios et al., 2018), excel in size and scope, although long-term access to them remains unreliable. Hand-annotated corpora, like the Moral Foundations Twitter Corpus (Hoover et al., 2020) and RedditBias (Barikeri et al., 2021), provide high levels of precision for the study of a particular phenomenon by supplying rigorous human-coded labels. Constructed bias-probing datasets, such as StereoSet (Nadeem et al., 2021) and BBQ (Parrish et al., 2022), offer carefully curated stimuli allowing researchers to isolate language-model biases. Longitudinal corpora like COHA (Davies, 2012) span multiple decades, thus allowing investigators to examine long-term linguistic shifts.

However, no existing resource, individually or collectively, satisfies all the key needs of computational research on organic social group discourse: (a) a naturalistic text source reflecting spontaneous communication; (b) a longitudinal setup; (c) the simultaneous inclusion of multiple, curated social group distinctions; (d) a multi-label framework combining both basic descriptors (like time and location) and more complex semantic variables (like emotion and moralization); (e) convergent validation against established measures; and (f) a multi-format architecture that accommodates projects of different scopes and levels of computational complexity. The Illinois Social Attitudes Aggregate Corpus (ISAAC) was developed to overcome these limitations. Table 1 situates ISAAC in this broader landscape, comparing it with existing resources in terms of size, time span, labels, and primary methodological constraints with respect to ISAAC's intended use.

**Table 1**

*Comparison of ISAAC to Existing Corpora*

| Corpus | Size | Years | Labels included | Distinctive limitation vs. ISAAC | Reference |
|---|---|---|---|---|---|
| Pushshift/Arctic-Shift | 5.6B+ comments by 2019 | 2005–2019; 2020–2025 | None (raw text only) | Unfiltered; no semantic labels; discontinued broad access | Baumgartner et al., 2020; Heitmann, 2026 |
| GoEmotions | 58K comments | Sample from 2005–2019 | 27 emotions (multi-label, single model) | Small; single-model labels; no target groups; no geography | Demszky et al., 2020 |
| Moral Foundations Reddit Corpus | 16K comments | Snapshot | 8 moral sentiment categories | Small; one construct; 12 subreddits | Trager et al., 2022 |

| Moral Foundations Twitter Corpus | 35K tweets | Snapshot | 10 moral sentiment categories (5 foundations × 2 valences) | Short text; 7 curated topic domains; no longitudinal axis | Hoover et al., 2020 |
|---|---|---|---|---|---|
| Social Bias Frames | 45K posts / 150K annotations | Re-annotation | Offensiveness, target group, implied statement | Curated from offensive sources; not population discourse | Sap et al., 2020 |
| RedditBias | ~12K annotated comments | Snapshot | Bias along gender, race, religion, queerness | Small; geared toward language model evaluation | Barikeri et al., 2021 |
| Civil Comments/Jigsaw | ~2M comments | 2015–2017 | Toxicity, identity mentions | Toxicity-framed; no complex semantic set | Borkan et al., 2019 |
| HolisticBias/BBQ/StereoSet/CrowS-Pairs | 1.5K – 450K | Constructed | Stereotype probes for language models | Templated or crowd-written, not naturalistic discourse | Nadeem et al., 2021; Nangia et al., 2020; Parrish et al., 2022; Smith et al., 2022 |
| Corpus of Historical American English (COHA) | ~475M words | 1820s–2010s | None (raw, register-balanced) | Edited registers, not lay attitudes; decade granularity | Davies, 2012 |
| Google Books Ngrams | 500B+ words | 1500–2019 | Ngram counts only | Different type of discourse; growing scientific text share in the 1900s (Pechenick et al., 2015) | Michel et al., 2011 |
| Twitter Decahose/TweetsKB | 3.1B tweets | 2013–2023 | Entity links, single-model sentiment | Decahose closed in 2023; geotags on <3% of tweets | Fafalios et al., 2018 |
| ISAAC (present paper) | 527M+ Reddit posts | 2007–2023 | Moralization, sentiment × 3, emotion ensemble × 3, generalization (clause-level), user localization, anonymization | — | — |

**The Illinois Social Attitudes Aggregate Corpus (ISAAC)**

The Illinois Social Attitudes Aggregate Corpus (ISAAC) is a corpus of 527,060,919 Reddit submissions and comments, posted between 2007 and 2023 and pre-filtered for relevance to six social group distinctions, including sexuality (straight versus gay), race (White versus Black), age (young versus old), ability (abled versus disabled), weight (thin versus fat), and skin tone (light-skinned versus dark-skinned). Each post is paired with a set of computationally generated labels. These labels include basic descriptors, such as (a) a randomly generated user ID, (b) time of posting, and (c) user location. More complex semantic labels include (a) moralization (whether a post frames its content in moral terms); (b) sentiment (the overall positive-to-negative evaluative tone of a post); (c) emotion (the extent to which a post expresses each of six basic emotions or affectively neutral content); and (d) linguistic generalization (how far a post's claims extend beyond particular individuals and episodes to the underlying categories).

The platform's capacity to support large-scale computational social science research on social group discourse rests on five core pillars.

First, the multi-stage filtering pipeline is conservative: stratified random samples of the final corpus contain less than 10% content judged irrelevant by trained human coders across all six social group distinctions and the full year range — an end-to-end benchmark that, to our knowledge, no comparable corpus of naturalistic social group discourse reports. The resulting low false positive rate enhances validity, reliability, and statistical power.

Second, ISAAC pairs naturalistic lay discourse at unusually large scale — two to four orders of magnitude larger than existing labeled comparators — with continuous 2007–2023 coverage under a consistent labeling pipeline. Data prior to 2007 were excluded due to Reddit's limited early scope; 2023 was chosen as the endpoint to avoid the confounding structural shifts in U.S. social media use that began in 2024 (Törnberg, 2026).

Third, ISAAC covers six social group distinctions in parallel through a single, unified pipeline and offers the broadest collection of labels of any comparable resource (see Table 1), including reliable estimates of users' home locations. Semantic labels cover constructs of long-standing interest in the computational study of social attitudes, with in-house modeling efforts concentrated where open resources are weakest or absent.

*Moralization* tracks whether a text frames its content in moral terms, such as judgments of right and wrong. Its volume and intensity have been a sustained concern of computational research (e.g., due to links with greater message diffusion; Brady et al., 2020). *Sentiment*, the most widely used text-derived discourse marker in the social sciences (Mäntylä et al., 2018; van Atteveldt et al., 2021), denotes the overall evaluative tone of a text and provides the most direct window onto valenced discourse about a group. Beyond broad valence, discrete *emotions* such as anger, fear, and joy (Ekman, 1992) carry distinct signatures in group-related language and serve as a central element in work on how attitude-relevant content spreads online (Berger & Milkman, 2012; Brady et al., 2020). Finally, statements differ in their level of linguistic *generalization* (e.g., "My gay neighbor kindly watered my plants while I was traveling" versus "Gay people are nice”). This distinction has been central to studying the persuasiveness of attitude-relevant communication (Kalla & Broockman, 2020; Thomas & Grigsby, 2024); yet no existing corpus of social group-related discourse offers generalization-focused labels.

Fourth, ISAAC serves users of varying computational experience, requiring no programming background for standard use: the core corpus and its stratified samples are accessible without code via https://isaac.psychology.illinois.edu/ and companion labeler web apps. Script-based dataset integration and single-line access modes are available to computationally trained researchers. To our knowledge, no comparable naturalistic corpus of social group discourse offers a similarly wide range of access pathways.

Fifth, ISAAC is designed as an open, modular corpus-construction pipeline rather than a single, static dataset. Because the underlying architecture is fully public, alternative source data, keyword sets, language filters, relevance classifiers, label resources, or location models can each be substituted.

Together, these design features position the corpus to support empirical research across multiple overlapping classes of questions. The open, modular architecture and diverse access pathways ensure that findings generated via ISAAC remain replicable and forward-compatible. The parallel coverage of six social group distinctions under a unified pipeline enables direct cross-category comparisons that single-topic corpora cannot accommodate. The 17 years of continuous coverage allows researchers to track long-term change and test event-locked hypotheses around cultural flashpoints without assembling a custom corpus for each event. Finally, the corpus’s location estimates connect discourse with important downstream measures and outcomes, including attitude and hate crime data.

Beyond these advantages, ISAAC is subject to three key limitations: (a) ISAAC targets aggregate linguistic trends rather than the micro-dynamics of interpersonal communication; (b) ISAAC labels observable language features rather than presupposing internal attitudes, leaving the text–attitude relationship for future investigation; and (c) ISAAC draws on a single platform, Reddit, with its own demographic and discursive idiosyncrasies, where limited expected future access to raw data may constrain temporal extensions. However, the open, modular design allows future investigators to overcome each of these limitations.

**Validation Strategy**

To ensure maximum data fidelity, ISAAC's validation strategy combines internal algorithmic verification with psychometric testing of construct validity.

We first verify the filtering and labeling pipelines both component by component and end to end. This internal validation is executed through systematic human audits of stratified data samples, evaluating the consistency and agreement among off-the-shelf ensemble labels, and calculating the classification accuracy and performance of in-house models on held-out validation data. We release the validation data alongside the corpus.

Then, we establish the construct and convergent validity of semantic and spatial indicators by mapping discourse volume and semantic corpus markers directly onto independent external benchmarks. This external validation framework relies on cross-referencing online discourse markers against four distinct categories of macro-societal data: (a) the timing of major cultural flashpoints, (b) the outcomes of historical state-level ballot initiatives, (c) independent internet search trends, and (d) national public survey data. These cross-methodological linkages demonstrate that the corpus effectively captures shifts in public attitudes across time and geography.

# Method

## Overview of Corpus Construction

The corpus was constructed in three phases (see Figure 1). First, a four-step filtering pipeline extracted social group-relevant discourse from a raw Reddit archive while aggressively removing irrelevant content.

**Figure 1**

*Overview of the ISAAC Construction and Access Pipeline*

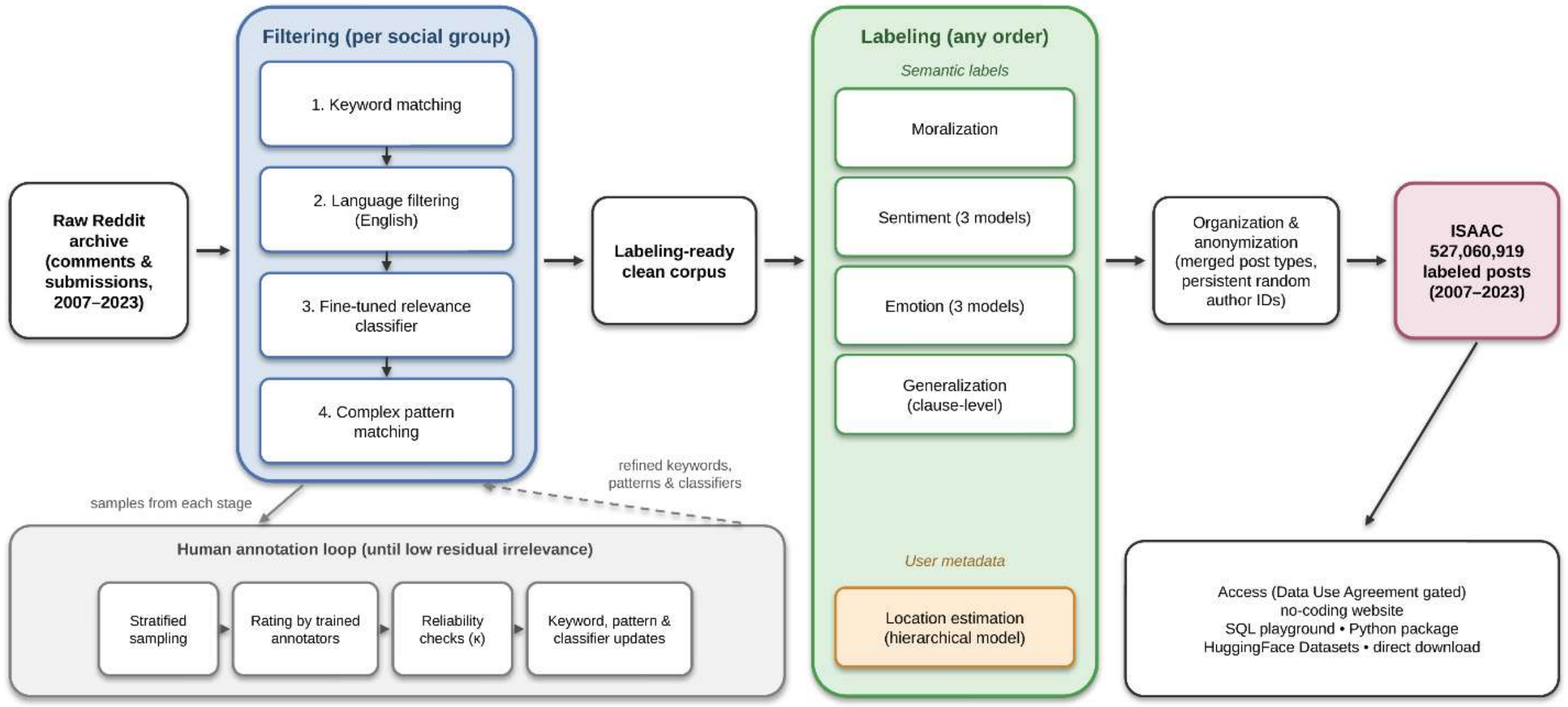

Along with each post, every ISAAC entry includes three sets of variables (see Table 2): (a) Reddit metadata, including unique post and parent identifiers, timestamps, and subreddit; (b) in-house user-level metadata, including a persistent, anonymized user identifier and a hierarchically organized estimate of the user's home location; and (c) semantic labels generated via in-house and off-the-shelf algorithms.

**Table 2**

*Variable Families in the Released Corpus*

| Variable family | Corpus columns | Values |
|---|---|---|
| Reddit metadata | Post and parent identifiers; text; creation time (GMT); subreddit; relative upvote/downvote score; post type (comment vs. submission); group-relevant matched patterns | Identifiers and free text; timestamp; integer (score); categorical (type); keyword list |
| User metadata | Anonymized user identifier; estimated home location with model confidence; second-most-likely location with confidence | Persistent random identifier; hierarchical location labels (U.S. states; non-U.S. regions: Africa, Americas, Asia-Oceania, Europe; unknown), confidence values between 0 and 1 |
| Moralization | Whether the post moralizes its content | Binary (0 = not moralized, 1 = moralized) |
| Sentiment (three models) | Counts of positive, neutral, and negative sentences (Stanza) | Sentence counts (non-negative integers) |
| | Average overall sentiment (VADER) | Compound polarity from -1 (negative) to 1 (positive) |
| | Average polarity and subjectivity (TextBlob) | Polarity from -1 (negative) to 1 (positive); subjectivity from 0 (objective) to 1 (subjective) |
| Emotion (three models) | Per-model scores for anger, disgust, fear, joy, sadness, surprise, and neutral content | Continuous scores from 0 to 1 per emotion and model |
| Generalization | List of clauses; per-clause generalization labels; counts and proportions of clauses by genericity, eventivity, boundedness, and habituality, plus clauses for which these features are undefined | Free text (clauses); categorical labels; counts (non-negative integers) and proportions (0 to 1) |

*Note.* The full column-by-column documentation is in the project repository (variable_list.md).

**Ethical Approval**

ISAAC's research design was reviewed by the University of Nebraska–Lincoln Institutional Review Board, which determined that it does not meet the regulatory definition of human-participant research under 45 CFR 46.102. Corpus development nevertheless followed current ethical recommendations for social media research (e.g., in anonymization protocols; Fiesler et al., 2020; Proferes et al., 2021). Details are provided in Appendix A.

**Data Source**

ISAAC was built from preexisting Pushshift-formatted comment and submission Reddit archives (Baumgartner et al., 2020; Heitmann, 2026) that have since been discontinued. Because we aimed for a comprehensive representation of discourse, we retained all content, did not filter the corpus for bots, and imposed no minimum post length. These decisions preserve the full range of naturalistic discourse for downstream researchers, who can apply stricter inclusion criteria suited to their own research questions.

**Computational Infrastructure**

Development was carried out both locally and on high-performance computing clusters. Early development relied on a cluster of four Nvidia A100 GPUs. The released corpus was constructed using the AI Innovation Institute cluster at Stony Brook University, with up to two dozen simultaneous task instances involving up to 200 CPU cores and 20 GPUs (comprising

H100 PCIe, Quadro RTX 8000, and Tesla V100-SXM2-32GB Nvidia units). An interactive web application hosted on the ISAAC website demonstrates the entire end-to-end pipeline using an illustrative subset of the corpus.

**Filtering**

To extract discourse relevant to the six social group distinctions from the raw archive, data were processed through a sequence of four filtering steps (see Figure 1), optimized to minimize the proportion of irrelevant posts retained in the final corpus.

***Step 1: Keyword Matching***

The first step scanned the raw Reddit archive using a keyword-based approach to comprehensively capture how social groups are discussed across the entire Reddit ecosystem. This method retains conversations wherever they occur, including within subreddits whose primary focus lies elsewhere, offering a broader representation than a restrictive, subreddit-based method. For example, while subreddit-based filtering limits data to specialized hubs like r/lgbt or r/BlackPeopleTwitter, keyword matching includes discourse in general communities, such as r/movies.

We used extensive keyword lists covering both poles of each social group distinction (e.g., "gay" and "straight" for sexuality). The lists were compiled through iterative searches of multiple thesauri and supplemented with Wikipedia's lists of slurs and group-specific slang. Review of stratified samples during corpus development led us to add terms that the initial lists missed and to remove noisy or outdated entries. Full keyword lists are available in the project's Git repository (https://github.com/BabakHemmatian/Illinois_Social_Attitudes). This step used exact matching of keywords over more complex pattern matching to remain computationally tractable given the volume of the platform archive (see algorithmic efficiency details in Appendix E.3).

### *Step 2: Language Filtering*

The second step retained only those posts identified as written in English by fastText (Joulin et al., 2017), an automated language-identification model that provides the high processing speed required to scan billions of posts. This step was necessary because Reddit includes substantial non-English content in primarily English-speaking communities, and many English keywords appear in a range of languages (e.g., “gay” in Filipino, Spanish, or French).

### *Step 3: Relevance Classification*

The third step removed irrelevant word meanings by applying a dedicated relevance classifier to each social group. This step was critical because initial keyword matching was optimized for comprehensiveness. As such, many posts were captured that used terms in ways that were unrelated to a social group distinction (e.g., “black” describing a physical color rather than a racial category). A separate relevance classifier was created for each social group distinction by fine-tuning the base RoBERTa language model (Liu et al., 2019) using domain-specific labeled sets. Because RoBERTa interprets words in the context of the full post, it can accurately distinguish intended from unintended uses of keywords in a way that static lists cannot.

Each classifier was trained on ~1,500 posts drawn from the keyword-filtered corpus, sampled randomly with stratifications for years and per-post keyword match counts. Each sample was rated for relevance to the target social group by two independent trained annotators. Annotator training involved discussing rating instructions (see Appendix B), followed by working through specific examples to ensure comprehension. For classifier training, unclear labels were marked as irrelevant, and residual disagreements were resolved in favor of relevance.

To ensure uniform data fidelity across ISAAC, the pipeline architecture enforced a conditional retraining protocol: when human audits of a social group distinction indicated that residual irrelevance remained above the 10% design threshold, classifiers were systematically retrained on additional rated samples drawn, and more complex pattern matching was applied. Retrained classifiers were tuned to be conservative, prioritizing the exclusion of borderline posts.

***Step 4: Complex Pattern Matching***

The fourth step applied social-group-specific sets of complex regular expression patterns to remove remaining irrelevant text. This step was necessary because the irrelevant content that survived relevance classification tended to fall into structural patterns that human readers easily recognize but that the classifiers continued to mislabel. These blind spots included proper names and pop culture references containing category-relevant words (such as “Walter White”).

Because the preceding steps had already reduced the corpus substantially, more computationally demanding pattern matching could be executed at scale. For instance, “black” would continue to be matched on its own but not as part of more complex but irrelevant phrases such as “black-and-white.” To maintain scalability, we used a fast pattern matching engine that we further parallelized (see Appendix E). The pattern sets were developed iteratively by inspecting stratified samples of surviving posts and are included in the project repository.

***Filtering Quality Assessment***

Filtering quality was monitored during development with an ongoing human annotation loop. Trained annotators rated stratified random samples after each processing step: (a) after initial relevance classification, (b) after the first application of complex pattern matching, (c) through subsequent pattern matching iterations (if applicable), and (d) after classifier retraining (if applicable). To ensure the reliability of human ratings, we double-annotated data samples at the beginning and the end of the filtering process for Reddit comments. Because filtering thresh-olds were developed and optimized using comment data, additional samples were drawn to confirm that filtering quality transferred to submissions. Development concluded once the residual irrelevance rate satisfied our target threshold of 10% across both post types and all social group distinctions.

## Metadata

### *Reddit Metadata*

Each ISAAC entry retains the Reddit metadata most useful for empirical research, including the unique identifiers for the post and (if applicable) its parent submission, creation time in GMT, subreddit, score (the relative count of upvotes and downvotes, with positive numbers indicating more upvotes), post type (submission versus comment), and the set of keywords that matched the post during filtering. The unique identifiers support thread reconstruction within the corpus and linkage of a user's posts across rows. To protect user privacy, using identifiers to re-link posts to identifiable Reddit accounts is prohibited (see Appendix A).

### *User Location Estimation*

Because no ground-truth location labels exist for Reddit users, location training labels were generated automatically from explicit textual self-disclosures (Harrigian, 2018). A model was then trained to estimate each user's home location from their full posting history, weighting three classes of signal: word usage, subreddit participation, and posting timestamps. The model is hierarchical: it first distinguishes U.S. from non-U.S. users, then assigns U.S. users to one of 51 state-level jurisdictions (the 50 states and the District of Columbia) and non-U.S. users to one of four world regions (Africa, Americas, Asia-Oceania, Europe). When the model's preference for the top label at a finer level is weak, the entry retains the coarser label, and entries with no sufficient support are marked unknown.

We report five kinds of validation evidence for location labels: (a) a manual audit of the automatically generated training labels; (b) classification performance on users never seen in training; (c) the same evaluation with explicit location mentions removed, showing level of reliance on overt self-disclosure; (d) calibration of the reported confidence values; and (e) the alignment of state-level volume surges around localized public votes on marriage equality. Methodological details of the model, its application to the corpus, and the calibration analyses are provided in Appendix D. Training data are not released due to user privacy concerns. The trained models are available by emailing isaac.corpus.support@gmail.com under the ISAAC Model Use Agreement, which prohibits attempts to re-identify individuals, bars surveillance, profiling, and other harmful applications, and requires citation of this article.

### *Anonymization*

After filtering and labeling comments and submissions separately, the comment and submission datasets were merged into a single time-ordered dataset per social group distinction. Usernames were then replaced with random identifiers. To preserve user-level dependencies and to allow for longitudinal and spatial analyses, each user retains the same identifier across all their posts and all six social group distinctions. To protect user privacy, the mapping between usernames and identifiers is not distributed with the corpus.

## Semantic Labels

### *Moralization*

We generated a binary moralization label by tuning a classifier on the Moral Foundations Reddit Corpus (MFRC; Trager et al., 2022). MFRC was chosen as the training source because it samples the same platform as ISAAC. Twitter-trained alternatives (e.g., MFTC; Hoover et al., 2020) are less well-suited for this pipeline due to differences in platform-specific language styles and conversational context (see Atari et al., 2023, for an alternative approach to adapting classifiers across platforms). For training, MFRC's annotations were reduced to moralized versus non-moralized, with disagreements resolved by majority vote and residual ties broken toward the moralized label to maximize sensitivity to moral content.

We supply a single binary label rather than labels for individual moral foundations for two reasons. First, published per-foundation classification performance is highly uneven, and applying unevenly performing labels across ISAAC's six social group distinctions would produce corresponding unevenness in downstream construct validity. Second, a binary construct commands broader theoretical agreement than specific foundations, around which debate continues.

The trained classifier is publicly released, and its performance on a held-out portion of MFRC is reported. To evaluate the validity of moralization classifications, we examine moralization distributions across the six social group distinctions and between their binary poles, with the prediction that highly contested social domains such as race will exhibit higher baseline moralization rates than less highly contested ones such as body weight.

***Sentiment***

Single-model sentiment labels are known to be noisy and to carry biases inherited from each model's training material (Chan et al., 2021). Therefore, we applied three established models with deliberately distinct designs: (a) VADER (Hutto & Gilbert, 2014), a rule-based model built specifically for social media text; (b) TextBlob (Loria, 2018), which derives continuous polarity and subjectivity scores from a dictionary of scored words developed against movie reviews (built on the Pattern lexicon; De Smedt & Daelemans, 2012); and (c) Stanza (Qi et al., 2020), a neural network model with broad training coverage that provides discrete sentence-level labels of positive, neutral, or negative. Each model's outputs are included in the final dataset as separate columns (see Table 2). Researchers can construct ensemble estimates (e.g., mean polarity or majority vote), restrict analyses to posts on which the models agree, or treat disagreement itself as a signal of interpretive difficulty.

The internal consistency of the sentiment ensemble was evaluated using a stratified random validation sample of the corpus ($k$ = 60,551 posts). This validation subset was constructed by sampling up to 500 comments and 100 submissions from a randomly selected monthly data archive for each individual social group distinction and calendar year, spanning systematically from 2007 to 2023. The final sample remained below the maximum target of 61,200 entries due to lower baseline data volumes within less populated social group distinctions during early years.

As a test of validity, we probe whether the evaluative tone about the marginalized poles of the social group distinctions (such as gay, Black, and fat people) leans more negative than the dominant poles of the same distinctions (such as straight, White, and thin people).

***Emotion***

Single-model emotion labels are even noisier than single-model sentiment labels, given uneven per-emotion performance on Reddit text and the relative scarcity of large, labeled training sets (Bostan & Klinger, 2018; Demszky et al., 2020). As such, we applied three emotion classifiers built on pre-trained language models: emotion-english-distilroberta-base (Hartmann, 2022), roberta-base-go_emotions (SamLowe, 2022), trained on the GoEmotions corpus of human-annotated Reddit comments (Demszky et al., 2020), and EmoBERTa (Kim & Vossen, 2021). The three classifiers were used jointly so that each of the six commonly used basic emotions (anger, disgust, fear, joy, sadness, and surprise; Ekman, 1992), along with neutral content, could receive three independent estimates per post.

Each model's scores, corresponding to one value from 0 to 1 per emotion per model, are included in the final dataset as separate columns. Scores from the first and third models are label probabilities that sum to one across the seven categories, whereas scores from roberta-base-go_emotions are independent per-category probabilities and need not sum to one. The model-specific ratings support the same ensemble and disagreement analyses as the sentiment columns.

We establish the reliability of the emotion labels by measuring cross-model agreement among the classifiers. We also probe the validity of emotion labels by evaluating the distribution of discrete emotions between category poles, expecting that discourse about marginalized social groups will be characterized by more negative emotions overall.

***Generalization***

Generalization was modeled as the combination of clause-level linguistic features (Hemmatian, 2022): (a) genericity (whether the clause's main referent is a generic category or specific individuals); (b) eventivity (whether the clause describes a stable state or a transient event); and, for eventive clauses, (c) boundedness and habituality (whether the events are temporally bounded or habitually recurring). An in-house classifier suite trained on a unique corpus (Hemmatian et al., in preparation) decomposed each post into clauses and labeled each clause on these features. Because the linguistic dimensions are undefined for non-statements (such as questions and imperatives), the pipeline filters and counts those instances separately.

Each entry in ISAAC carries the list of its clauses, the per-clause labels, and per-feature counts and proportions (see Table 2). This clause-level resolution allows researchers to trace how gradations of anecdotal focus and broad generalizations are interwoven within a single post. For each feature, we report held-out classification performance; the underlying annotated corpus and training procedure are documented in the companion paper (Hemmatian et al., in preparation).

**Corpus Access**

ISAAC is distributed through five access pathways tailored to different research scales and workflows. The first route is optimized for easy, coding-free access, whereas the four remaining pathways are designed to support programmatic or large-scale composable workflows.

First, a coding-free interface hosted at https://isaac.psychology.illinois.edu/ allows registered users to download the corpus, subsets, or reproducible stratified random samples. The platform delivers outputs in CSV format (see Figure 2). Second, an SQL query playground on the website supports limited-scale filtered and aggregated queries against the corpus without large downloads. Third, the isaac-data Python package, installable from the standard PyPI repository, reads the corpus's download catalog and loads selected social group distinctions, year ranges, and columns directly into standard data analysis workflows, with resumable downloads for offline work. Fourth, HuggingFace Datasets and Models Hub interfaces support seamless incorporation into machine learning and AI pipelines. Fifth, instructions for direct download allow registered users to script custom data retrieval pipelines. Details and links are in Appendix F.

**Figure 2**

*Screenshot of the Coding-Free ISAAC Website*

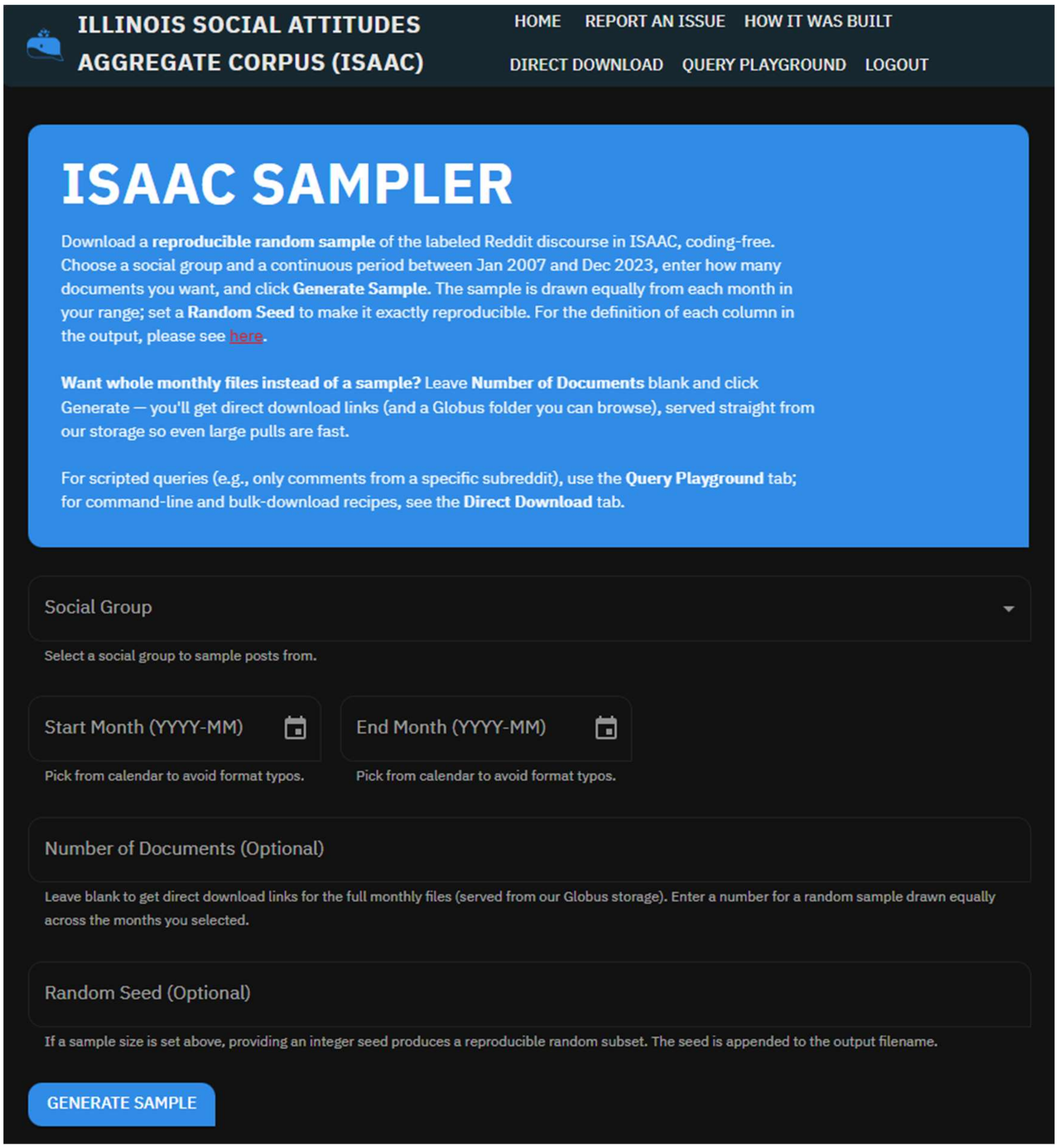

*Note.* The website is hosted at https://isaac.psychology.illinois.edu/.

**Data Use Agreement**

All use of ISAAC's data, tools, and scripts through the different access pathways is governed by a Data Use Agreement developed in consultation with the Office of Legal Counsel at the University of Illinois Urbana–Champaign (see Appendix A). Access to ISAAC is restricted to non-commercial academic research, with prohibitions on user re-identification, generative AI training, and limits on redistribution and copies. The agreement bans use to harm, discriminate against, or surveil individuals or groups of individuals. All downstream researchers are required to use official access channels.

**Adapting and Extending ISAAC**

To support projects with different data workflows and levels of computational complexity, corpus development scripts are public, documented, and modular. This architecture allows other teams to seamlessly apply the same operational steps to new datasets or to adapt the pipeline to new use cases.

For researchers who wish to apply ISAAC's in-house models to their own texts without writing code, a web application hosted at https://huggingface.co/spaces/BabakScrapes/isaac-classifiers provides access to the relevance classifiers (all six social group distinctions behind a single selector), the moralization labeler, and the generalization suite (clause segmentation, per-clause labels, and summary proportions). Each classifier accepts both single texts and uploaded files containing multiple texts. The applications draw on shared, dynamically allocated fast processing capacity and are therefore subject to per-user quotas and queueing, making them suited to exploratory use and moderate volumes; for large-scale labeling, we recommend downloading the tool and using the pipeline's hardware-acceleration support. The underlying models carry the licenses described in the Code Availability statement. Use of the applications is additionally governed by the hosting platform's acceptable-use policy.

Because all resources, configuration defaults, and trained models are publicly released, every step of ISAAC's construction is transparent and auditable, and the pipeline can be applied wholesale to any data source that follows the same schema. With broad public distribution of the raw Reddit archives upon which ISAAC drew discontinued, the released corpus also serves as a durable point of access to this slice of the platform's history. Adaptation to other targets requires changing only the component that differs from ISAAC's design: (a) novel social group distinctions or even entirely different topics can be captured by substituting keyword lists; (b) non-English corpora can be produced by changing the target language label; (c) custom relevance classifiers and location models can be fit on user-supplied data using the companion training resources; (d) new human annotations can be processed and verified using the stratified sampling and interrater agreement tools; and (e) non-Reddit data sources can be ingested by modifying only the data-reading utilities of the first filtering step. Details are provided in Appendix E.

**Analytic Strategy**

Model performance of the filtering, location, and semantic models is evaluated using point estimates derived from held-out or manually audited validation samples. Corpus-level analyses show census descriptives of the entire dataset; because no sampling is involved, no confidence intervals are reported. For analyses relying on independent units or stratified subsamples, we report exact tests, native metrics (such as $\kappa$, $F_1$, $\rho$, and $r$), and bootstrapped 95% confidence intervals where appropriate.

For multi-class evaluations (e.g., when evaluating the performance of the location model), we focus on three standard machine learning measures: precision (how often an assigned label is correct), recall (how much of each true category the model captures), and the $F_1$ score, the harmonic mean of the two. Values closer to 1 indicate better performance. We rely on macro-averaged $F_1$ scores as the primary metric to ensure a conservative, class-balanced assessment. Macro-averaging computes performance per label and weights each category equally, preventing high-performing, heavily populated categories from artificially masking poor performance in rarer or less populated categories. Because this work focuses on estimation and validation rather than hypothesis testing, no power analyses were conducted.

To evaluate the macro-level validity of the finalized dataset, the corpus-level analytic strategy implements three distinct validation checks centering on text volume and semantic characteristics. First, macro-level convergent validity was evaluated by correlating corpus volume fluctuations against Google Trends search patterns. Second, external validity was assessed by comparing monthly post volumes around social group distinction-specific major anchor events against pre-event baselines. To establish the validity of the location labels, this event analysis was complemented by evaluating localized discourse spikes surrounding nine state-level marriage equality ballot measures spanning 2008 to 2012 against non-voting state baselines. Semantic labels were validated by examining cross-label interactions and correlating a longitudinal composite sentiment score against Gallup's annual marriage equality public opinion tracking data (Gallup, 2026).

## Results

### Filtering Pipeline Performance

The corpus development target was less than 10% irrelevant content in year- and keyword count-stratified random samples of comments for every social group distinction. Once human rater reliability of the training samples was established (see Appendix C.1) and the target threshold was reached, a final step ensured that high filtering quality transferred from comments to submissions.

The number of steps to achieve the 10% target varied by social group distinction. For age, sexuality, and body weight, the goal was reached after a single neural network-based classi-fication and complex pattern matching sequence. Managing complex contextual noise in the ability, race, and skin tone domains required a multi-stage pipeline, including reinforced classification and secondary pattern matching passes. Relevance classifier performance is reported in Appendix C.2. Step-by-step error reduction tables tracking each sequential pruning checkpoint are provided in Appendix C.3.

Interrater agreement on the final audit samples was high for each social group distinction, with raw agreements ranging from 95.3% to 99.0% and Cohen's κ ranging from .556 to .852. For age, κ is uninformative due to the classic prevalence problem (Feinstein & Cicchetti, 1990): because only one or two sampled documents were irrelevant, the baseline for expected chance agreement approaches 100%, artificially suppressing the metric despite near-perfect raw accuracy.

Across double-rated final stratified samples spanning all social group distinctions, the residual irrelevance rate was 3.3% (23/700) under a lenient rule where a single rater's approval suffices, and 6.1% (43/700) under a stringent rule where a post counts as relevant only if both independent raters agree (see Table 3). The discrepancy between the two rules is typical for social media text: it closely mirrors the human disagreement rates reported for comparable constructs, such as Social Bias Frames (Sap et al., 2020) and the Moral Foundations Reddit Corpus (Trager et al., 2022).

**Table 3**

*Final Residual Irrelevance Audit on Comments and Submissions*

| Distinction | Final stage | *k* | Comments rate (stringent rule) | Comments rate (lenient rule) | κ | Raw agreement | Submissions rate (single-rated) |
|---|---|---|---|---|---|---|---|
| Ability | Submissions pattern pass | 100 | 6.0% | 3.0% | .651 | 97.0% | 4.0% |
| Age | Comments pattern pass | 100 | 2.0% | 0.0% | .000 | 98.0% | 2.0% |
| Body weight | Comments pattern pass | 100 | 4.0% | 3.0% | .852 | 99.0% | 2.0% |
| Race | Submissions pattern pass | 150 | 8.0% | 5.3% | .786 | 97.3% | 4.0% |
| Sexuality | Comments pattern pass | 100 | 5.0% | 2.0% | .556 | 97.0% | 5.0% |
| Skin tone | Submissions pattern pass | 150 | 9.3% | 4.7% | .643 | 95.3% | 9.0% |
| Pooled | — | 700 | 6.1% | 3.3% | — | — | 4.3% |

*Note.* Under the stringent rule a post counts as relevant only if both raters agree; under the lenient rule either rater suffices. All columns except the last are based on double-rated comment samples; the final column is a single-rater transfer audit of submissions ($k$ = 100 per distinction; pooled $k$ = 600). Stage labels give the last filtering step each distinction required to reach the development target (see Appendix C).

## Key Descriptive Statistics

Of the 527,060,919 entries in the final corpus, 64,581,610 (12.25%) were Reddit submissions and 462,479,309 (87.75%) Reddit comments. The volume varied considerably by social group distinction, with age being the most discussed category, followed by race, sexuality, skin tone, ability, and body weight (see Figure 3). The number of entries per year increased for all social groups across successive collection years (see Figure 4).

**Figure 3**

*Number of Entries in ISAAC per Social Group Distinction*

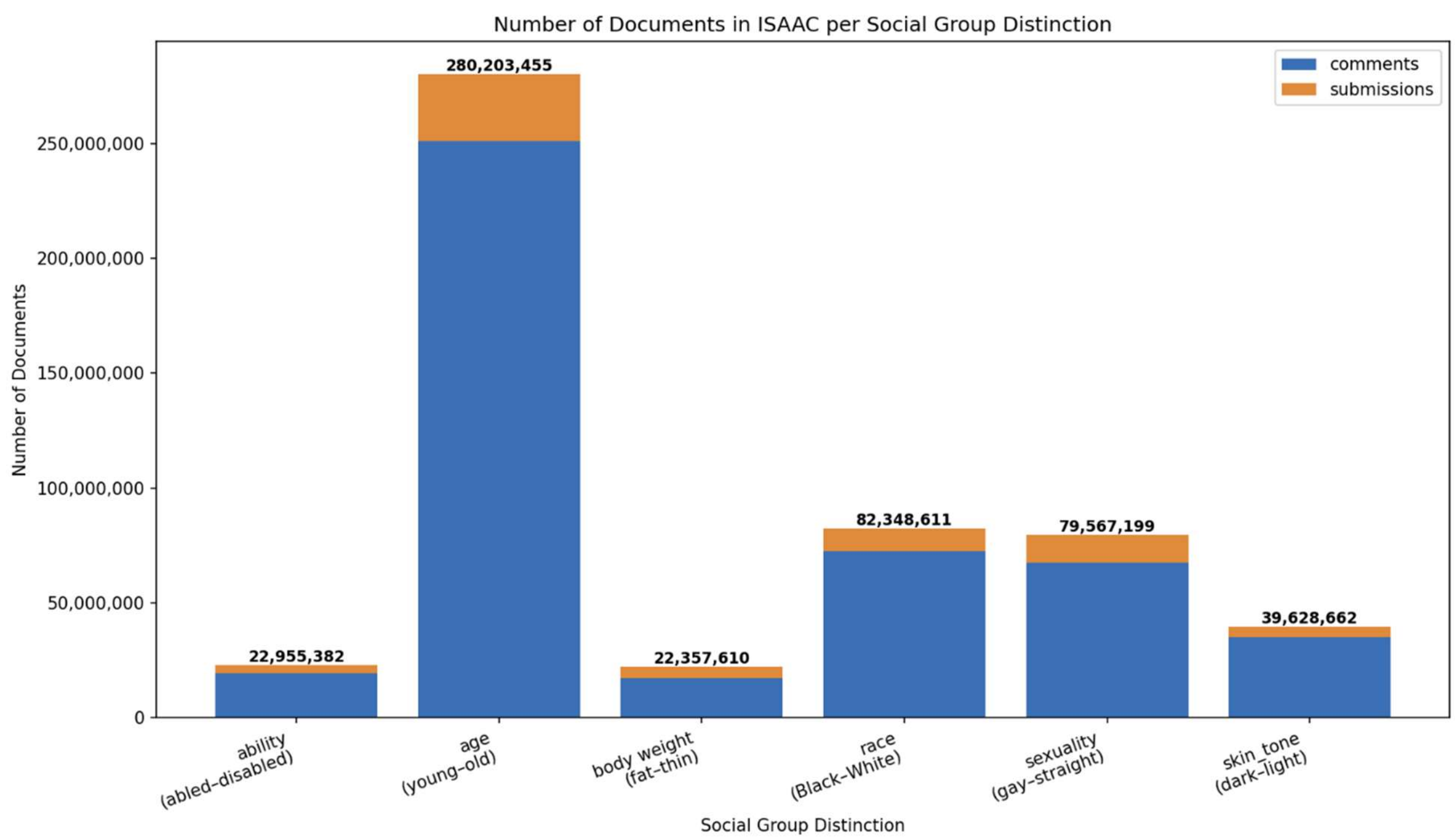

*Note.* Binary poles are given in parentheses.

**Figure 4**

*Posts per Social Group Distinction Over Time, Pooled across Comments and Submissions*

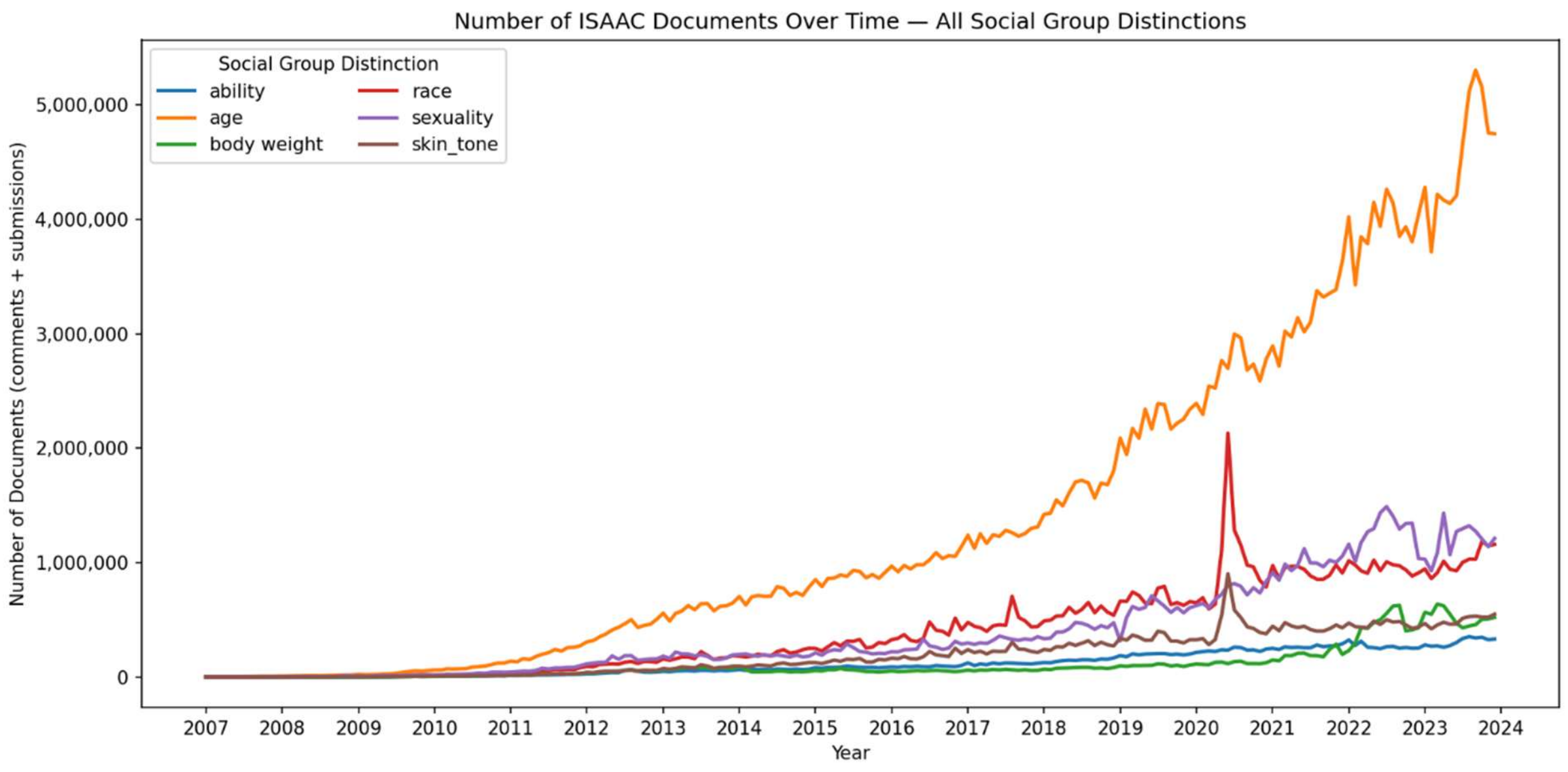

## Location Model Evaluation

Comprehensive details about the location model development, inference, and evaluation are reported in Appendix D. We highlight the key results in this section.

To ensure the integrity of the auto-labeled data used to train the location model, a manual audit validated the accuracy of the text-extracted self-disclosures against available post context. The extraction pipeline demonstrated high data fidelity across all geographic tiers, achieving 95.7% accuracy at the macro U.S. versus non-U.S. level (88/92), 90.2% at the U.S. state level (46/51), and 90.0% at the non-U.S. global region level (36/40). These audit figures confirm the reliability of the training data used to fit the location model.

To evaluate how well the deployed hierarchical model generalizes to unseen data, we tested it on a set of 11,875 held-out users (6,766 U.S.-based; 5,109 non-U.S.-based). The model successfully separated U.S. from non-U.S. users (macro $F_1$ = .913; raw accuracy = .915). Among resolvable non-U.S. users ($n$ = 5,102), the four-global-regions model achieved a macro $F_1$ of .825 and raw accuracy of .881. For the 51-class U.S. state model (50 states plus the District of Columbia; *n* = 6,413), the ensemble reached a macro $F_1$ of .698 and raw accuracy of .754.

The errors at the state level were relatively minor: the correct state appeared within the top five predictions for 91.1% of users, the median distance error between the predicted and true state centroids was 0 km, and 75.6% of all predictions fell within 100 km of the true geographic centroid (see Table 4).

Labeled users all self-identified their locations in their posting history. To assess how well the model performs on users in the broader corpus who may not do the same, we re-evaluated the model's performance after stripping city, state, and country names from the validation users' texts. This masking process left coarse geographic classification largely intact (pre- vs. post masking macro $F_1$ of .913 to .900 for U.S. vs. non-U.S. and .825 to .789 for world region), but U.S. state-level performance dropped sharply (from .698 to .371). The top-k and within-radius accuracy sweeps in Table 4 confirm that even when the model missed the exact state under masking, it tended to place the user in a nearby state or in the correct U.S. census region. Together, these results suggest that the model's errors are spatially structured, thus making ISAAC well-suited for discourse marker research questions at the state–year level.

Nonetheless, to ensure that higher resolution labels were assigned only when the user's self-identification or distinctive language use warranted them, conservative confidence thresholds across the full corpus restricted assignments to coarser geographic tiers when no clear winner emerged at the finer level. Most of ISAAC's 30,098,293 unique authors received a distinct location label under this scheme. Non-US world region designations were dominated by Europe. The relative shares of specific US states (comprising ~5% of total users) were highly consistent with the 2024 census (Pearson's r = .93, .88 when weighted by post frequency) despite systematic deviations (see Appendix D.3.5 for the detailed distributions).

The location and the runner-up designation for each labeled user are accompanied by scores quantifying model uncertainty. At the U.S./non-U.S. and global region tiers, the labeler was consistently underconfident, but its rank ordering identified the more reliable estimates. State-level scores are included for completeness but are unreliable: the tiered thresholds already removed ambiguous cases, leaving accuracy roughly uniform across assigned labels at the finest level. Appendix D (Sections 3.6–3.7) reports detailed calibration analyses and the recalibration mappings released with the corpus.

**Table 4**

*Held-Out Location-Model Performance Under Standard and Masked Conditions*

| Metric | Level | n | Standard | Masked | Δ |
|---|---|---|---|---|---|
| Macro $F_1$ (top-1 choice) | U.S. vs. non-U.S. | 11,875 | .913 | .900 | −.013 |
| Macro $F_1$ (top-1 choice) | Non-U.S. region (4) | 5,102 | .825 | .789 | −.036 |
| Macro $F_1$ (top-1 choice) | U.S. state (51) | 6,413 | .698 | .371 | −.327 |
| Accuracy (top-1 choice) | U.S. state | 6,413 | .754 | .476 | −.278 |
| Accurate label among top-5 | U.S. state | 6,413 | .911 | .754 | −.157 |
| Accurate label among top-10 | U.S. state | 6,413 | .943 | .836 | −.107 |
| Median centroid error (km) | U.S. state | 6,413 | 0 | 211 | +211 |
| Within 100 km | U.S. state | 6,413 | .756 | .479 | −.277 |
| Within 500 km | U.S. state | 6,413 | .803 | .607 | −.196 |

*Note.* Distances use U.S. Census state centroids. Calibration and recalibration of the reported confidence values are reported in detail in Appendix D (see Figure D3).

**Semantic Label Validation and Performance**

Table 5 outlines the evaluation framework applied to the semantic labels across two distinct blocks. For integrated, off-the-shelf sentiment and emotion tools, we report ensemble agreement metrics. Details of the creators' validation benchmarks are available in the respective publications and summarized in the table. For custom, in-house models, we report empirical performance on held-out validation slices.

***Moralization***

The custom moralization classifier was evaluated on a held-out 10% slice of the Moral Foundations Reddit Corpus (MFRC; $k$ = 2,682; 49.1% moralized). For the moralized class, the model achieved a precision of .733, a recall of .788, and an $F_1$ score of .760, with an overall accuracy and cross-class macro $F_1$ of .755. These metrics confirm that the classifier captures moralized framing in line with expectations for automated subjective discourse analysis.

***Sentiment***

The two lexicon-based models (VADER and TextBlob) demonstrated stronger mutual agreement than either did with the neural network architecture (Stanza; see Figure 5a). One source of divergence is Stanza's pronounced negative skew: it classifies 64.0% of posts as negative (VADER: 35.1%; TextBlob: 20.3%). Adjusting for each model's baseline scoring tendencies and re-evaluating their category assignments yielded only a minor performance improvement, confirming that the models rank individual posts differently beneath the aggregate skew. Nevertheless, at the ensemble level, the final three-model average exhibited moderate reliability (ICC = .63). These figures align with the baseline limitations of the domain: trained human annotators frequently achieve only moderate agreement on social media sentiment indicators ($\alpha \approx .6$; Mozetič et al., 2016).

### *Emotion*

Pairwise categorical agreement across included models on top emotion labels was modest, with Cohen's κs ranging from .01 to .33 (see Figure 5b). The primary reason was highly divergent baseline rates. For instance, the GoEmotions model assigned a neutral label to 83% of posts, reflecting its neutral-heavy Reddit training distribution. Equalizing detection rates improved pairwise categorical alignment, with κs between .09 and .43. The roberta-base-go_emotions and emotion-english-distilroberta-base models yielded the closest overall match, with κs ranging from .30 to .43 across all categories.

Continuous intensity scores demonstrated substantially higher alignment than top labels: the three-model ICC ranged from .53 to .68 across all categories, indicating moderate-to-good reliability for ensemble-level intensity tracking across all seven emotions. When scaled across the entire corpus, majority voting successfully resolved these individual model biases to assign a clear two-of-three consensus emotion label to 77.7% of all posts, yielding roughly 410 million reliably categorized records. Unanimous agreement was achieved in 24.6% of cases.

### *Generalization*

The underlying corpus and segmentation/labeling model metrics are detailed in a companion paper (Hemmatian et al., in preparation). The automated segmentation tool underlying the labels covered most of the target clause span for 95.5% of human-verified clauses. On a 10% held-out slice of the training corpus, the full 18-way situation entity classifier that identifies different mixtures of target linguistic features achieved an accuracy of .737, with heavy label imbal-ance constraining its cross-class macro $F_1$ to .514. Collapsed to the three linguistic features incorporated into ISAAC, $F_1$ was .852 for genericity (accuracy = .860), .879 for eventivity (accuracy = .894), and .804 for the four-class boundedness/habituality feature (accuracy = .850).

**Table 5**

*Label-Classifier Performance*

| Source | Construct | Model | Evaluation set | *k* | Performance |
|---|---|---|---|---|---|
| **Off-the-Shelf (Published Benchmarks)** | Sentiment | VADER | Tweets vs. human raters | 4,000 | $F_1$ .96 (3-class); $r$ = .881 with human intensity |
| | | TextBlob/Pattern | Movie reviews | — | ≈ 75% polarity accuracy |
| | | Stanza | SST 3-class test (+4 corpora) | — | 70.0% accuracy |
| | Emotion | emotion-english-distilroberta-base | Balanced 6-corpus held-out | ~4,000 | 66% accuracy (7-class; chance 14%) |
| | | roberta-base-go_emotions | GoEmotions test split | ~5,400 | Macro $F_1$ .54 (28-label multi-label, threshold-optimized) |
| | | EmoBERTa (base) | MELD / IEMOCAP | — | Weighted $F_1$ 65.61 / 67.42 |
| **In-House Models (New Held-Out Evaluations)** | Moralization | BERT (base) | MFRC 10% held-out | 2,682 | Moralized class P .733 / R .788 / $F_1$ .760; macro $F_1$ .755; accuracy .755 |
| | Generalization | 18-way situation entity | Gold corpus 10% held-out | ~2,357 | Macro $F_1$ .514; accuracy .737 |
| | | Genericity | Gold corpus 10% held-out | ~2,357 | Macro $F_1$ .852; accuracy .860 |
| | | Eventivity | Gold corpus 10% held-out | ~2,357 | Macro $F_1$ .879; accuracy .894 |

| Source | Construct | Model | Evaluation set | *k* | Performance |
|---|---|---|---|---|---|
| | | Boundedness/ha-bituality | Gold corpus 10% held-out | ~2,357 | Macro $F_1$ .804; accuracy .850 |

*Note.* Off-the-shelf rows summarize the evaluations published by each model's creators on their own benchmarks; in-house rows report new held-out evaluations conducted for ISAAC.

**Figure 5**

*Ensemble Internal Agreement on the Stratified Subsample*

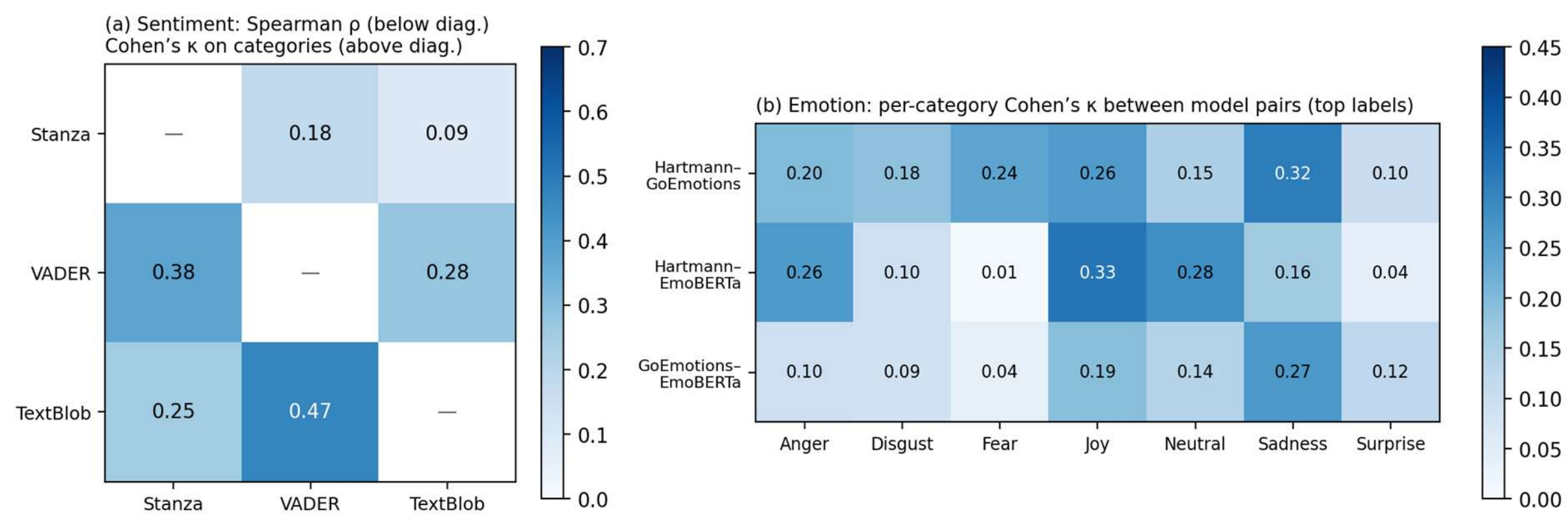

*Note.* $k$ = 60,551. (a) Sentiment: pairwise Spearman ρ on continuous outputs (below diagonal) and Cohen's κ on categorical codes (above diagonal). (b) Emotion: per-category Cohen's κ on top labels for each model pair; all three members score the same seven categories.

## Corpus-Level Validation

### *Volume-Based Validation*

**Search Interest.** Monthly ISAAC text volumes demonstrated strong baseline alignment with independent Google Trends data across the 17-year tracking window, with Spearman's ρ reaching .96 for skin tone, .88 for ability, .82 for race, .79 for weight, and .33 for age (see Figure 6). Sexuality was the sole exception, yielding a negative uncorrected baseline correlation of $\rho = -.68$. However, when normalized as a proportional share of total platform traffic, the corpus volume and search interest data aligned positively, $\rho = +.63$. This reversal confirms that the initial negative correlation was an artifact of conflicting platform trends: raw search interest receded after marriage equality legal milestones, whereas raw Reddit volume expanded along with general platform growth.

To isolate sudden temporal shifts from long-term longitudinal growth trends, we evaluated the first differences of monthly volumes across the timeline. Month-over-month volume changes remained consistently positive for all social group distinctions, $.03 \leq \rho \leq .35$. This alignment suggests that both the platform archive and external search engine patterns actively register shared, short-term public discourse events at a highly compressed monthly timescale.

**Figure 6**

*Convergent Validity of Discourse Volume Against Search Interest*

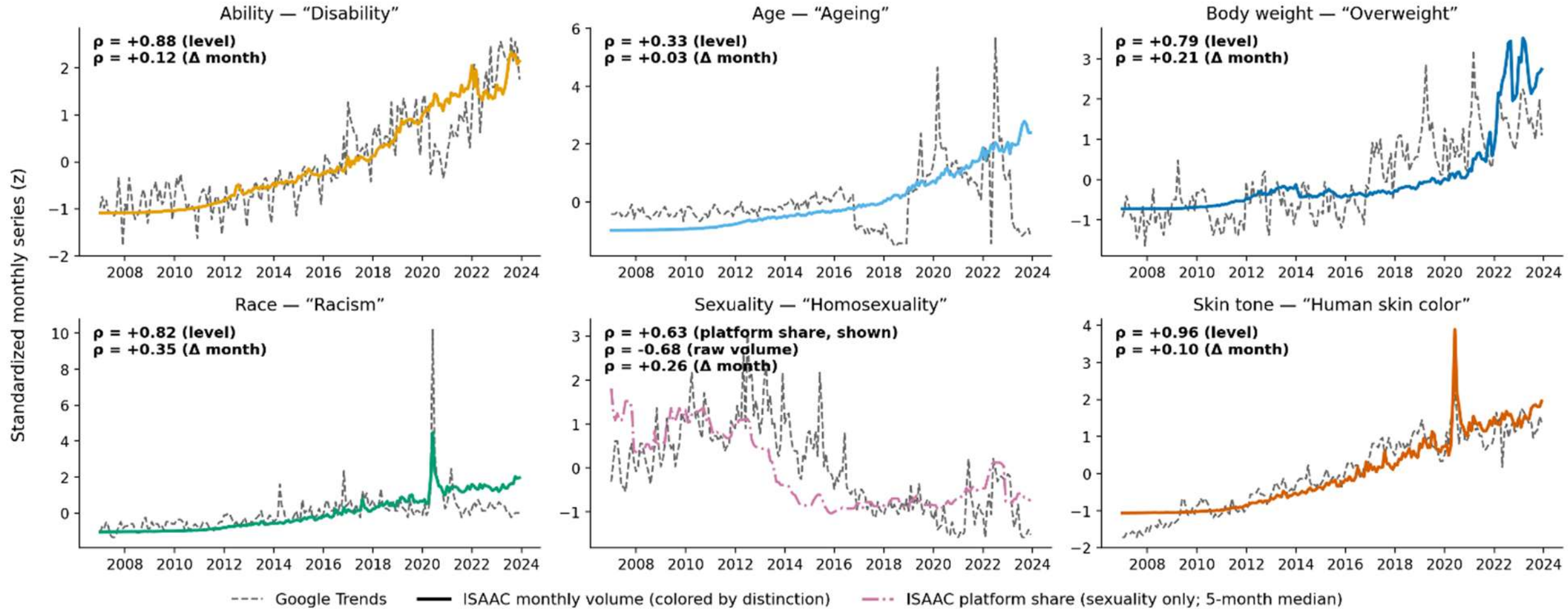

*Note.* Standardized monthly ISAAC volume (solid) is plotted against Google Trends search interest for the matching topic (dashed), 2007–2023. Annotations give Spearman ρ at the level and first-difference (Δ month) grains. For sexuality, the ISAAC series is instead shown volume-adjusted (dash-dot).

**National Event Spikes.** To evaluate ISAAC's responsiveness to acute external shocks, group-specific posting volumes around major societal events were compared against trailing baseline windows (see Figure 7). To account for platform growth, we report each event month's percentile relative to all post-2013 months, along with its share-adjusted volume relative to total platform traffic.

Sexuality-related discourse experienced a sharp spike during the June 2015 *Obergefell v. Hodges* Supreme Court ruling, with volume rising 46% above its baseline, landing in the 98$^{th}$ percentile of all post-2013 months. The volume of race-related postings increased substantially in mid-2020 following the murder of George Floyd and the ensuing Black Lives Matter resurgence, rising 65% in May and a massive 219% in June, marking the single largest month in the social group distinction's 17-year record (99.2nd percentile; +153% share-adjusted). Skin tone postings surged concurrently with the racial justice movement in June 2020 (+173%, 99th percentile), an alignment consistent with colorism discourse moving in close tandem with broader national racial dialogues. Ability-related discourse expanded significantly during the January 2017 Affordable Care Act repeal debate when disability coverage centered the national policy fight, running 32% above its baseline (99th percentile) and surviving platform-growth adjustments (+12% share-adjusted, 97th percentile).

Body weight discourse underwent the largest sustained shift in the repository; following Wegovy's FDA shortage listing in March 2022, volume climbed to +157% above baseline by that summer, remained +138% elevated through December 2023, and peaked in March 2023 at +194% raw and +224% share-adjusted, demonstrating the GLP-1 public attention wave. Conversely, age serves as an informative null result, rising a modest 12% in raw volume at the onset of COVID-19 in March 2020 but falling 4% below baseline when share-adjusted for platform growth (23rd percentile). Because age represents ISAAC's broadest and largest topic — spanning diverse themes such as parenting, generational divides, and aging bodies — the sheer variety of its coverage naturally dilutes event-locking for any single occurrence (see Charlesworth et al., 2023, for a similar dissociation between age and other distinctions in a large social group attitudes dataset).

**Figure 7**

*Event-Locked Posting Volume by Social Group Distinction*

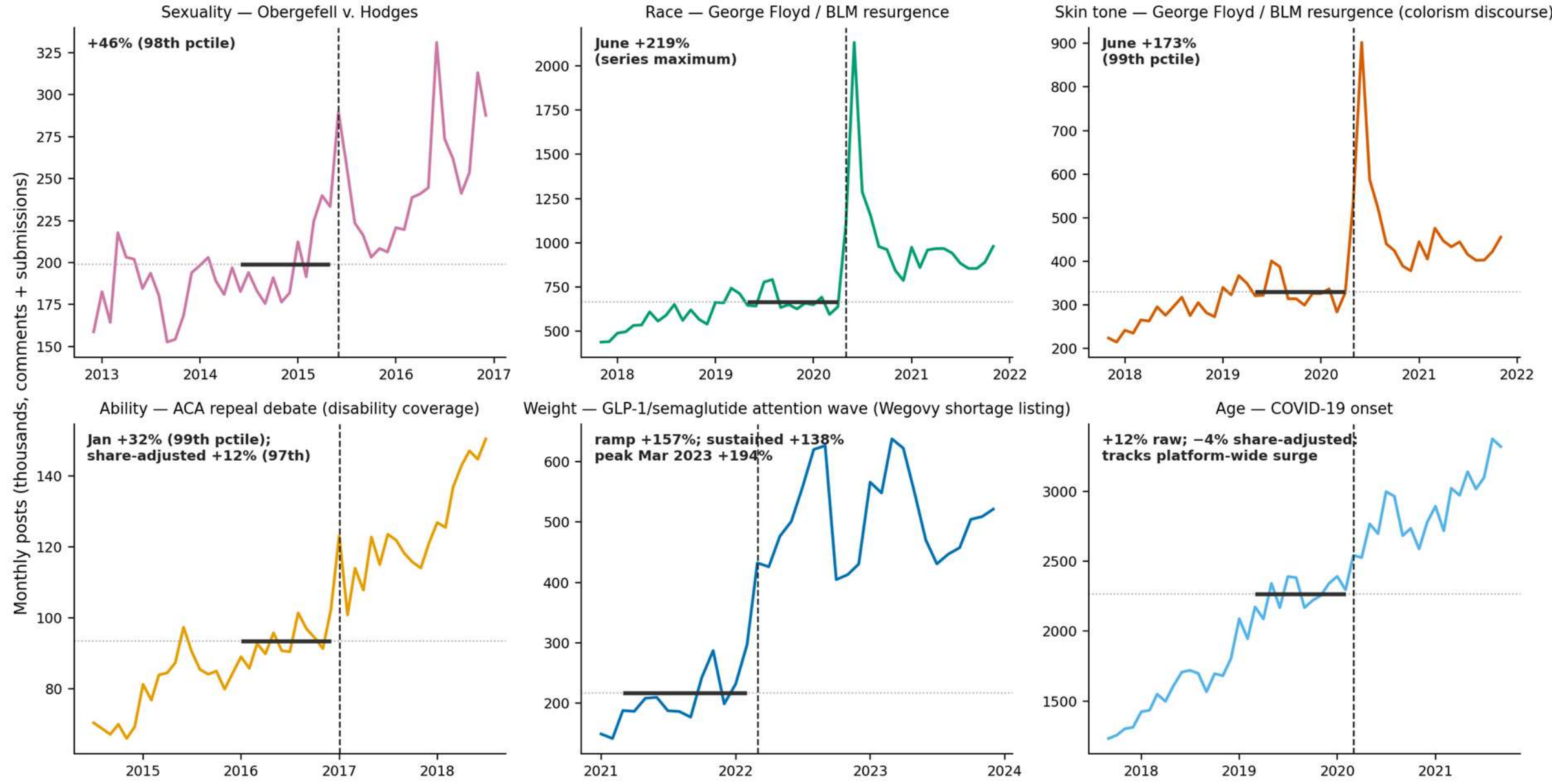

*Note.* Pooled comments and submissions. Dashed vertical lines mark event months; the horizontal bar marks each distinction's baseline window. Percentages give proportional change against baseline; percentiles locate the event month among all post-2013 months on the same statistic.

**Localized Event Spikes.** To validate the location labels, we examined sexuality discourse volume surrounding the nine state-level same-sex marriage public votes that occurred within ISAAC's 2007–2023 window. For each election, we compared sexuality-relevant posting volume by users whose estimated home state was holding the vote against the corresponding index for all non-voting states (see Appendix C.4 for the full procedure).

Sexuality-related discourse rose sharply within voting states around each of the nine marriage equality ballot initiatives, reaching more than eight times the state's own trailing baseline in California and North Carolina. Because national attention to votes could inflate discourse everywhere, we benchmarked each voting state's elevation against the same index computed for non-voting states at the same dates. Voting states exceeded the non-voting median in all nine cases, by factors ranging from 1.07 to 3.85 (see Figure 8). This 9-out-of-9 outperformance was statistically significant under an exact two-sided sign test, $p = .004$. A case-level bootstrap placed the geometric mean elevation at 1.78, 95% CI [1.39, 2.34]. The spikes occurred regardless of whether the ballot initiative aimed to restrict or extend marriage equality.

**Figure 8**

*Location Validation via Marriage-Equality Votes*

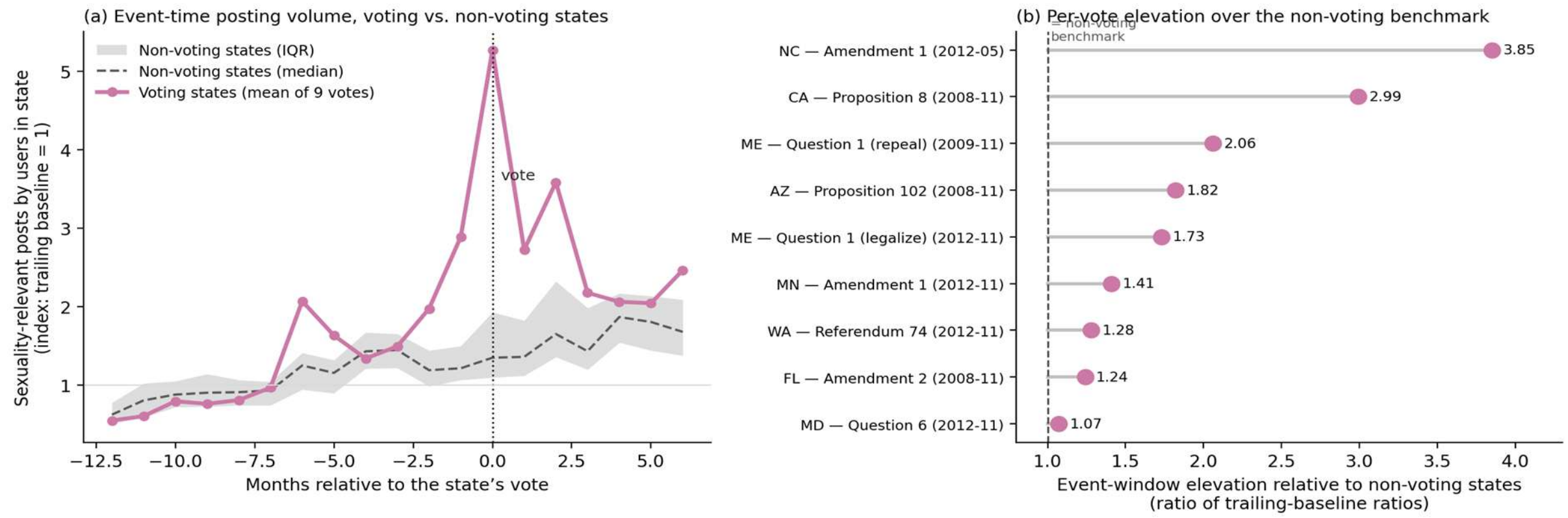

*Note.* (a) Sexuality-relevant posting volume by users in voting states (mean across the nine votes), aligned to the vote month ($t = 0$) and indexed to each state's trailing baseline (months −14 to −3), against the median and interquartile range of the identical index for non-voting states (baseline ≥ 5 posts/month) at the same calendar dates. (b) Each vote's event-window elevation relative to the non-voting benchmark; values above 1 indicate state-specific elevation beyond national salience. Maine's pre-vote spike in panel (a) reflects its May 2009 legislative legalization, which triggered the veto referendum.

### *Semantic Validation*

**Inter-Variable Alignment.** Across volume-weighted corpus samples, moralization rates remained high across all six social group domains, ranging from 49% to 74% of all posts (see Appendix C.5 for the detailed profiles across social group distinctions and a comparison with the rates reported by Atari et al., 2023). What differentiated social group domains from each other was not the presence of moralized rhetoric itself, but rather the unique affective and emotional profiles that the discourse assumed within each group.

In 100,000-post volume-weighted samples for each social group distinction in the final corpus, discussions surrounding race and ability emerged as the most heavily moralized and negative. However, the two targets diverged sharply in their underlying emotional composition. Race was heavily anger-dominant (26.6% of posts), plausibly reflecting intense political and social conflict, whereas ability featured distinctively high rates of sadness (14.2%) and fear (3.9%), potentially characteristic of hardship- and illness-centered discourse. Conversely, sexuality remained the most neutral-heavy domain (58.4%), while body weight skewed highly positive as the least moralized (49.3%) and most joyful (28.7%) social group distinction, a pattern suggesting that supportive, self-improvement communities may dominate its corpus volume.

As a test of convergent validity, we examined the co-occurrence patterns between moralized framing and discrete emotional expressions. Across every social group distinction, moralized posts exhibited a pronounced surge in anger compared to non-moralized text. This signature moral outrage coupling was most acute in highly contested domains such as skin tone (+22 percentage points) and race (+19 percentage points; Brady et al., 2020). In contrast, the introduction of moralization into weight-related discourse was associated with a 15-percentage point decline in joy, capturing the linguistic footprint of fat shaming moralization cutting against an otherwise supportive baseline.

Pole-level sentiment contrasts confirmed that the marginalized identity pole consistently attracted more negative sentiment than its dominant counterpart across race, ability, sexuality, and body weight (see Table 6). This directional pattern reversed for the age and skin tone distinctions, where the old and dark-skinned poles tracked with more positive sentiment. The mechanisms driving this directional reversal remain unclear and thus present a target for future empirical investigation.

**Table 6**

*Label Distributions by Social Group Distinction and Pole*

| Distinction | Moralized (%) | Sentiment (*z*) | Negative emotion (%) | Sentiment, marginalized pole | Sentiment, dominant pole |
|---|---|---|---|---|---|
| Ability | 72.2 [71.3, 73.1] | −.11 [−.13, −.10] | 36.0 | −.14 | +.01 |
| Age | 67.1 [66.2, 68.1] | +.04 [+.03, +.06] | 37.2 | +.09 | +.03 |
| Body weight | 60.2 [59.2, 61.2] | +.11 [+.10, +.13] | 24.8 | +.02 | +.15 |
| Race | 68.1 [67.2, 68.9] | −.12 [−.14, −.11] | 35.5 | −.21 | −.04 |
| Sexuality | 69.5 [68.6, 70.4] | +.12 [+.11, +.14] | 30.6 | +.10 | +.22 |
| Skin tone | 67.5 [66.6, 68.4] | −.04 [−.05, −.02] | 32.8 | +.07 | −.06 |

*Note.* Stratified sample, $k = 60{,}551$, with year-equal weighting and bootstrap 95% CIs in brackets. Sentiment in ensemble $z$-units. Negative emotion share: mean of the three models' top-label indicators over anger, disgust, fear, and sadness combined. Pole assignment from matched keyword patterns; posts matching both poles excluded from pole columns.

**Validation Against National Public Opinion.** The annual composite tracking the sentiment of sexuality discourse in ISAAC exhibited strong long-term alignment with independent U.S. public opinion across the 17-year historical tracking window (see Figure 9). Specifically, the annual corpus sentiment composite (*k* = 101,888; see Appendix C.4 for its construction) was highly and significantly correlated with Gallup's national support readings for marriage equality, $r = .71$, 95% CI [.35, .89], $\rho = .69$, $p = .001$. This longitudinal alignment tracked macro shifts in public attitudes, with sexuality discourse recording its most negative valence in 2008, corresponding with Gallup's baseline support low of 40%, and its most positive valence in 2023 as national support plateaued at 71%.

Year-over-year directional increments did not systematically co-move, first-difference $r = -.08$, 95% CI [-.55, .43], indicating that the observed correspondence reflects a shared, decade-scale macro trajectory rather than more fine-grained temporal patterns. Gallup's steady upward trajectory beginning in 2009 preceded (rather than followed) the sharpest corpus sentiment gains recorded between 2018 and 2023.

**Figure 9**

*Sexuality Discourse Tone Against U.S. Same-Sex-Marriage Attitudes*

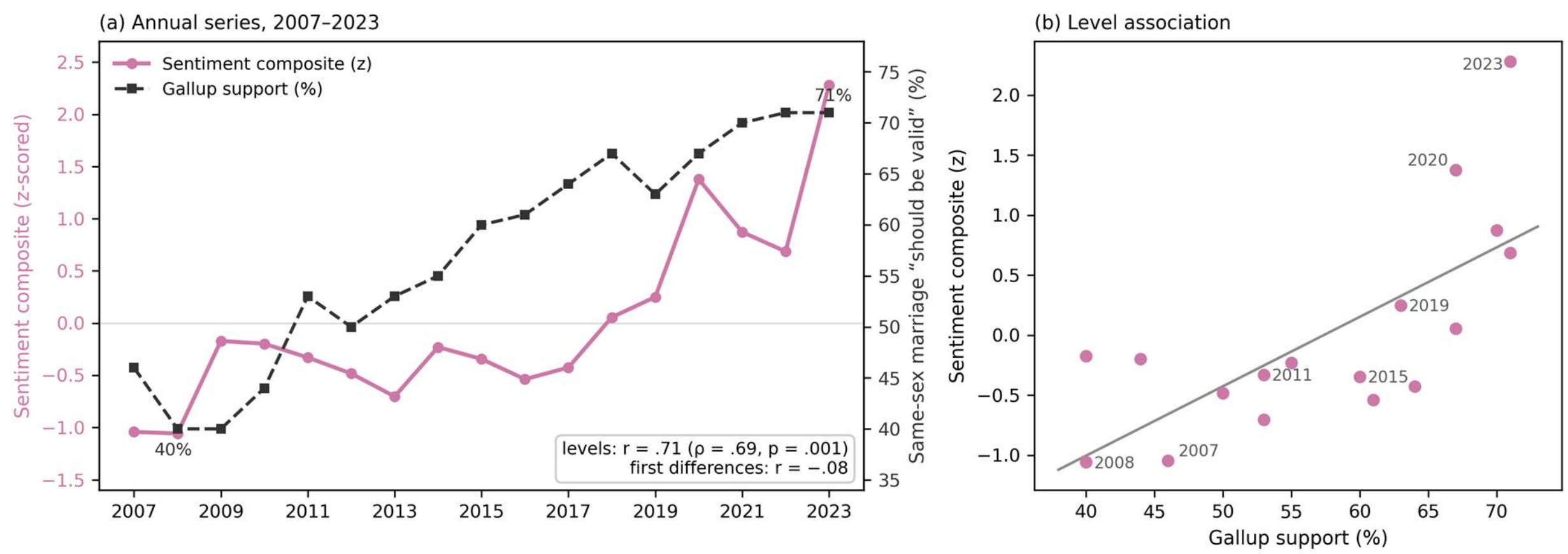

*Note.* (a) Annual three-tool sentiment composite for sexuality discourse (solid; mean of $z$-scored VADER, TextBlob, and Stanza scores over 500 comments drawn from every month of 2007–2023, $k = 101{,}888$) and Gallup's May reading of the share saying same-sex marriages should be legally valid (dashed, right axis; Gallup, 2026). (b) The same 17 annual pairs with a least-squares fit.

## Discussion

ISAAC was built to facilitate the at-scale study of public discourse about social groups without sacrificing measurement quality or flexibility. Each methodological choice that guided ISAAC's development was derived from that single goal. The corpus features more than 527 million Reddit posts, spanning 17 years and six social group distinctions of longstanding scientific interest, alongside three families of variables: platform metadata, hierarchically organized estimates of each user's home location, and semantic labels for moralization, sentiment, emotions, and generalization. Filtering was aggressive and multi-stage because topical noise compounds at this scale; location assignment was conservative and tiered because regional analyses inherit the error of their least certain labels; and semantic labeling was multi-model to provide better coverage of complex, subjective semantic constructs by avoiding single-classifier bias.

Validity evidence spans each level of the pipeline architecture. At the component level, human annotators agreed substantially to almost perfectly on the relevance of content to the target social group distinctions. Residual irrelevance, as rated by human coders in stratified samples, met the single-digit development criterion for every social group distinction and for both Reddit submissions and comments. The location model separated U.S. users, non-U.S. global regions, and U.S. states (where extractable) with high sensitivity and specificity. Finally, the semantic classifiers performed at or near state-of-the-art levels.

At the corpus level, macro-level volume and sentiment indicators aligned with external benchmarks. Specifically, ISAAC's volume converged with independent online search interest for each social group distinction and reflected major societal events both nationally and locally; moreover, a sentiment composite derived from the sexuality-relevant posts tracked gold-standard, longitudinal survey data across its historical window. Because no single comparison is definitive on its own, the core argument for the validity of ISAAC relies on convergence across multiple indices.

**ISAAC's Scope of Intended Use**

Within the growing ecosystem of open resources on social group attitudes, ISAAC occupies a unique niche. Whereas survey series — such as the American National Election Studies (ANES, 2026) or the General Social Survey (GSS, 2026) — feature representative self-report data, and repositories of indirect measures contribute large-scale data on implicit evaluations both in the United States (Nosek et al., 2007; Ratliff et al., 2020) and internationally (Charlesworth et al., 2023), ISAAC provides unobtrusive, naturalistic text data along with temporal, geographic, and semantic indicators. The corpus can be deployed independently or combined with these existing resources, allowing investigators to triangulate across divergent methodological approaches.

Beyond facilitating cross-methodological triangulation, ISAAC supports a cumulative approach to computational social science. Historically, the reliance on localized data collection and unique preprocessing choices has confined discoveries regarding social group-related public discourse to isolated, study-specific silos. Providing a fixed, public framework allows subsequent studies to directly build upon, extend, or contest existing findings within a unified architecture.

This structural continuity ensures that empirical insights can systematically accumulate over time rather than resetting with each new data collection and data analysis effort.

The parallel architecture of the pipeline also enables direct cross-category comparisons that are methodologically intractable using single-topic datasets. Processing multiple social group distinctions under an identical data harvesting and modeling framework ensures that variation observed across categories reflects genuine discursive patterns rather than pipeline artifacts. For example, while baseline moralization rates remained uniformly high across all six social group distinctions, the underlying affective profiles diverged sharply, separating the anger-dominant rhetoric of race from the sadness- and fear-heavy discourse of ability. This standardized comparative capacity allows researchers to isolate both universal and category-specific features of social group discourse at scale.

The temporal density of ISAAC enables the application of time-series and quasi-experimental frameworks to model both short- and long-term discourse dynamics. On a compressed monthly timescale, researchers can deploy interrupted time-series and difference-in-differences designs to isolate the immediate effects triggered by acute societal shocks, expanding on the national and localized event-locked analyses reported above. Across the full 17-year window, this continuous tracking supports lead-lag analyses capable of determining whether aggregate discourse features serve as a leading or lagging indicator of gradual, structural shifts in public opinion. Because the underlying corpus metrics systematically align with independent online search trends and national public tracking data, investigators can deploy these more sophisticated analytical frameworks confident that the text-derived signal is stable enough to support higher-resolution temporal modeling without confounding from internal tool drift.

The location labels provided with ISAAC support spatial designs that map digital discourse onto localized benchmarks, reflecting a broader shift toward tracking spatial variation across social and cognitive processes (Calanchini et al., 2022; Murphy et al., 2018). Because these location estimates reflect the linguistic footprint of consequential regional events — as demonstrated by the localized discourse surges surrounding state-level marriage equality ballots — ISAAC facilitates research investigating the relationship between text features and regional variables. Crucially, by anchoring these features to specific state–year boundaries, this spatial resolution extends the convergent validation framework pioneered by Caliskan et al. (2017) and Garg et al. (2018) from static, population-level word embeddings toward spatially patterned and potentially dynamic, longitudinal indicators. This advancement enables text-derived markers to be systematically linked to (a) regional attitudinal data, including state-level survey series and implicit evaluation repositories (e.g., ANES, 2026; Charlesworth et al., 2023; GSS, 2026; Nosek et al., 2007; Ratliff et al., 2020), and (b) external regional indicators, such as aggregate group-level outcomes, mobilization patterns, or legislative activity (Müller & Schwarz, 2021; Olteanu et al., 2019; Steinert-Threlkeld et al., 2015).

The semantic labels within ISAAC can be used to conduct large-scale, language-based tests of theories across the quantitative social sciences. For instance, the systematic coupling of moralized framing and affective outrage observed across the data links directly to established accounts of online communication dynamics (Brady et al., 2020), while the linguistic generalization metrics enable the systematic study of category-level generic and essentialist claims (Haslam & Whelan, 2008). Furthermore, for computational researchers, the corpus serves as naturalistic training and evaluation material for text classification models applied to complex, unstructured social discourse. The disaggregated multi-label architecture provided with ISAAC allows investigators to isolate distinct linguistic and semantic dimensions without relying on oversimplified composite metrics.

**Limitations and Possibilities for Expansion**

ISAAC is subject to several structural limitations, which affect any downstream use. First, the source data (a) reflect the demographic characteristics of the host platform, which skews predominantly young, male, and United States-based (Amaya et al., 2021), (b) are restricted to English-language content, and (c) are influenced by specific platform architecture norms (Tufekci, 2014). Consequently, the dataset indexes public discourse within this specific digital ecosystem rather than representative general population baselines. Second, the corpus contains disaggregated discourse markers rather than direct, latent attitude labels. Applications that use these manifest variables to make inferences about underlying attitudes require separate construct validation to account for the inferential gap between naturalistic text features and unobservable latent constructs (AlDayel & Magdy, 2021). Such uses of ISAAC will likely benefit from considering how evolving platform dynamics and digital affordances — such as systemic changes in user baseline volume over time, content moderation policies, and shifting algorithmic visibility — may shape manifest text expressions independently of underlying user attitudes.

Two additional structural attributes represent deliberate design compromises with documented mitigations. First, state-level geographic classification is restricted to approximately 5% of total users to minimize false positive contamination within regional sub-corpora. The inclusion of independent model confidence scores and recalibration mappings allows researchers to customize their own precision–coverage trade-offs based on specific design requirements. Second, cross-model classification agreement varies across the included sentiment and emotion estimators. Rather than obscuring these discrepancies within a single composite score, the repository publishes each model's independent output side by side with cross-model alignment statistics, rendering variability transparent and analytically tractable for downstream use.

The architectural modularity of the data pipeline supports systematic expansion across four distinct dimensions. Independent teams can extend the framework by (a) substituting keyword dictionaries to add novel social groups, (b) modifying data-loading components to accommodate text files from alternative digital platforms, (c) integrating multilingual classification filters to produce non-English corpora, or (d) appending additional custom semantic labels onto the database. Each expansion pathway is operationalized within the released open-source repository (see Appendix E). This modular design ensures that subsequent technical contributions from independent laboratories can systematically accumulate on a common, versioned infrastructure foundation.

## Conclusion

ISAAC facilitates cumulative quantitative social science by offering an integrated infrastructure for the study of naturalistic social group discourse at scale. Toward this end, the multi-decade text corpus comes with temporal, geographic, and semantic indicators across six distinct social group categories, derived using a standardized pipeline. To maximize downstream utility, ISAAC is distributed through multiple access pathways tailored to varied statistical or engineering requirements, while its modular, open architecture supports systematic expansion to overcome current limitations.

## Acknowledgments

We thank Eleanor Ruby Klein, Siyu He and Lauren Casey for assistance with human validation, Shantanu Sunil Dhamdhere and Vedant Mahajan for help with website development, and the Office of Legal Counsel at the University of Illinois Urbana–Champaign for consultations on the Data Use Agreement. Ty Villaneuva contributed logo and website design.

# Declarations

## Funding

This work was supported by startup funds provided by the University of Illinois Urbana–Champaign to Benedek Kurdi.

## Competing Interests

The authors have no relevant financial or non-financial interests to disclose.

## Ethics Approval

The research design was reviewed by the University of Nebraska–Lincoln Institutional Review Board (Project ID UNL-00024699; Form ID UNL-00064753), which determined on June 16, 2025 that the project does not meet the regulatory definition of human-participant research under 45 CFR 46.102. The Institutional Review Boards of the University of Illinois Urbana–Champaign and Stony Brook University deferred review to the University of Nebraska–Lincoln. Corpus development nevertheless followed current ethical recommendations for social media research (Fiesler et al., 2020; Proferes et al., 2021); full documentation is provided in Appendix A.

## Consent to Participate

Not applicable. The corpus is derived entirely from secondary analysis of publicly archived Reddit content and involves no intervention or interaction with living individuals.

## Consent for Publication

Not applicable. No identifying details of individual users are published; usernames are replaced with persistent random identifiers before release, and the mapping is not distributed.

**Data Availability**

The corpus generated during the current study is available at https://isaac.psychology.illinois.edu/, and through the isaac-data Python package (PyPI), a gated HuggingFace Datasets mirror, and direct download endpoints documented in Appendix F. One component is exempt: the training data underlying the location model are not publicly available because the training labels are derived from explicit textual self-disclosures of location and their release would permit cross-referencing to identifiable Reddit accounts, undermining the anonymization protocol. Access to data on every route requires acceptance of the project Data Use Agreement. The human relevance and quality-assurance ratings distributed in the pipeline repository are governed by the same agreement.

**Code Availability**

The corpus construction pipeline (all resource scripts, the command-line interface, keyword lists, regular expression pattern sets, and default configurations) is available under the MIT License at https://github.com/BabakHemmatian/Illinois_Social_Attitudes. The access website's frontend and backend are available under the same license at https://github.com/BabakHemmatian/ISAAC_Sampler and https://github.com/BabakHemmatian/ISAAC_Sampler_Backend.

Model weights are distributed through HuggingFace. Instructions for downloads are found in the main corpus repository. The six relevance classifiers, the moralization classifier, and the generalization suite are released under a Creative Commons Attribution 4.0 International License without access restrictions. The location model is released under a separate Model Use Agreement prohibiting re-identification, surveillance or otherwise harmful applications; access may be requested by emailing isaac.corpus.support@gmail.com.

**Authors' Contributions**

Babak Hemmatian: Conceptualization, Methodology, Software, Data curation, Validation, Formal analysis, Visualization, Investigation, Writing – original draft, Writing – review & editing, Supervision, Project Administration. Sarah Hadjarab: Software, Validation, Investigation, Writing – review & editing. Jessica Chen: Software, Validation, Writing – review & editing. Benedek Kurdi: Conceptualization, Validation, Writing – review & editing, Supervision, Funding acquisition, Resources.

**Open Practices Statement**

The full ISAAC corpus is openly available at https://isaac.psychology.illinois.edu/ and via HuggingFace Datasets, subject to the project Data Use Agreement. The complete corpus construction pipeline, including all keyword lists and pattern sets, is released under the MIT License at https://github.com/BabakHemmatian/Illinois_Social_Attitudes, documenting every processing decision from raw archive to released corpus. Relevance, moralization, and generalization models are distributed through HuggingFace under a Creative Commons Attribution 4.0 International License; the location model is available upon request under the ISAAC Model Use Agreement, as described in the Code Availability statement. Because broad public distribution of the source archives has since been discontinued, the released corpus, its stratified samples, and the accompanying validation data serve as the reproducibility anchor for the results reported here. The location model training data are withheld for user privacy reasons set out in Appendix A; the confidence recalibration mappings fitted on those data are released with the corpus. None of the analyses were preregistered.

# Appendix A

## Ethics and Data Handling Protocols

ISAAC's research design was reviewed by the University of Nebraska–Lincoln (UNL) Institutional Review Board (the first author's primary affiliation at the time), which determined on June 16, 2025 that the project (Title: The Discourse Predictors of Changes in Bias Towards Social Groups; Project ID UNL-00024699) does not meet the regulatory definition of human-subjects research under 45 CFR 46.102, on the grounds that (a) it is a systematic investigation of secondary, public data rather than research involving intervention or interaction with living individuals, and (b) user-level identifiers are removed before any analytic data use. The University of Illinois Urbana–Champaign IRB, home institution of co-investigator Benedek Kurdi, and Stony Brook University, the current affiliation of Babak Hemmatian, deferred review to UNL.

Notwithstanding the non-human-participants determination, we treat Reddit discourse about social groups as ethically sensitive even when technically public, and ISAAC's pipeline implements current social media research ethical recommendations (Fiesler et al., 2020; Proferes et al., 2021). Any use of the project's data, code repositories, or associated tools is governed by a Data Use Agreement developed in consultation with the Office of Legal Counsel at the University of Illinois Urbana–Champaign (version 2026-08-02; see here). Here, we highlight a few key points:

(a) the dataset is intended only for non-commercial research use;

(b) the organize_anonymize resource replaces Reddit usernames with persistent random identifiers as the final preprocessing step before release, so that the same user retains the same identifier across all their posts and across all six social distinctions;

(c) the anonymization mapping is held internally and is not distributed alongside the corpus, so published ISAAC rows cannot trivially be linked back to specific Reddit accounts;

(d) the source comment, parent, and submission identifiers are retained in ISAAC rows to support thread reconstruction, provenance auditing, and linkage to other Reddit-derived resources. This treatment is consistent with the norm for comparable open Reddit corpora (e.g., GoEmotions, Demszky et al., 2020; MFRC, Trager et al., 2022), which retain source identifiers for the same research-utility reasons;

(e) because the retained identifiers can in principle be joined to Reddit archives to recover the original Reddit username and full thread context, we do not claim that re-identification is technically prevented. We mitigate this residual risk by (i) not releasing the anonymization mapping between original usernames and ISAAC's persistent random user IDs; (ii) governing all use of the corpus through Data Use Agreement terms that prohibit re-identification, re-distribution and merging with Reddit archives for that purpose; and (iii) withholding the location training data from release entirely, since its generation relies on explicit textual mentions of geographic terms, including the text of the posts containing them, so that releasing it would in principle permit cross-referencing back to identifiable Reddit accounts. The trained location models do not carry this risk and are publicly released, as are the three train_location_* resources, so that interested researchers may train new location models on their own data.

The dataset website at https://isaac.psychology.illinois.edu, hosted on a university server at the University of Illinois Urbana–Champaign, requires users to register with an email address and to scroll through the then-current full Data Use Agreement before accepting it; the accepted version and timestamp are recorded server-side, and only the anonymized corpus is presented through this interface. The Python data-loader requires acceptance before any data access, re-prompts whenever the agreement text changes, and records each acceptance server-side with the user's email address. The dataset loader hosted on the HuggingFace Hub requests and documents agreement before granting access. Companion labeler demonstrations hosted on HuggingFace Spaces follow the platform's use policy.

# Appendix B

## Relevance Annotation Instructions

Our goal is to develop an algorithm that, given any text, can determine if its content is related to [social distinction] ([privileged pole] vs. [underprivileged pole]). Separately, we will train an algorithm that would identify the kind of attitude expressed towards [privileged pole] or [underprivileged pole] people. To help the development of these algorithms, we need related labels based on your intuition for the attached set of Reddit posts. Some of these texts will be unrelated to the topic, others will be related, and the relevance of some might be unclear. You can use the guide below to determine the labels for various situations accordingly.

[social distinction]_relevance:

0: Does not explicitly refer to or imply something about one or more individual or [social distinction] group presented as [privileged pole] or [underprivileged pole] even in passing

1: Explicitly refers to or implies something about one or more individual or [social distinction] group presented as [privileged pole] or [underprivileged pole] and their associated topics, even in passing x: Unclear association with someone or a group's [social distinction] attributes presented as [privileged pole] or [underprivileged pole], perhaps because of missed conversational context or incoherence

# Appendix C

## Supplemental Validation Details

### C.1 Annotator Agreement on Classifier Training Samples

Human annotations of stratified random samples underlying the relevance filtering pipeline achieved high interrater reliability across all six categories (see Table C1), with ability, sexuality, and age falling in Landis and Koch's (1977) "almost perfect" band ($\kappa \geq .81$), and race, skin tone, and weight (.690) falling in the "substantial" band (.61–.80). Weight's lower κ was driven by a base-rate difference between raters rather than disagreement on clear cases, consistent with its high raw agreement rate (89.6%).

**Table C1**

*Interrater Agreement on the Double-Rated Relevance Training Samples*

| Distinction | N | Raw agreement | Cohen's κ |
|---|---|---|---|
| Ability | 1,476 | 98.2% | .960 |
| Age | 1,485 | 90.6% | .812 |
| Body weight | 1,475 | 89.6% | .690 |
| Race | 1,498 | 92.6% | .784 |
| Sexuality | 1,493 | 93.6% | .871 |
| Skin tone | 1,885 | 93.4% | .725 |

*Note.* Two trained annotators rated the sample for each social group distinction.

### C.2 Relevance Classifier Performance

Table C2 reports precision, recall, and $F_1$ scores for the six relevance classifiers evaluated against held-out validation slices of the human-annotated training samples. The finalized classifier pipeline demonstrated high classification accuracy across all social group targets, yielding $F_1$ scores between .781 (skin tone) and .987 (body weight). Despite strong held-out performance of the corresponding models, iterative validation samples of race and skin tone continued to produce a long tail of complex false positives that resisted filtering. Reflecting the conditional optimization protocol, we retrained the associated models using targeted optimization samples and deployed a conservative classification threshold, after which the development target for false positives in stratified random samples was reached (see Table C2).

**Table C2**

*Relevance Classifier Performance on Held-Out Slices for Each Social Group Distinction*

| **Distinction** | **Classifier** | **Test set** | ***k*** | **Precision** | **Recall** | **$F_1$** |
|---|---|---|---|---|---|---|
| Ability | Initial | Initial 10% held-out | 148 | .815 | .880 | .846 |
| Age | Initial | Initial 10% held-out | 149 | .853 | .921 | .886 |
| Body weight | Initial | Initial 10% held-out | 148 | .975 | 1.000 | .987 |
| Race | Initial | Initial 10% held-out | 150 | .811 | .811 | .811 |
| Race | Retrained | Retraining 10% held-out | 40 | .786 | .917 | .846 |
| Race | Retrained | Initial 10% held-out | 150 | 1.000 | .757 | .862 |
| Sexuality | Initial | Initial 10% held-out | 150 | .962 | .949 | .955 |
| Skin tone | Initial | Initial 10% held-out | 189 | .862 | .714 | .781 |
| Skin tone | Retrained | Retraining 10% held-out | 40 | .875 | .840 | .857 |
| Skin tone | Retrained | Initial 10% held-out | 189 | .839 | .743 | .788 |

*Note.* Retrained race and skin tone classifiers use an asymmetric false positive penalty and a P(relevant) > .6 decision threshold; the released corpus uses the retrained checkpoints.

## C.3 Residual Irrelevance Across Development Stages

Figure C1 tracks residual irrelevance in stratified development samples across the successive filtering stages described in the main text, and Table C3 reports the corresponding rates at each checkpoint.

**Figure C1**

*Residual Irrelevance Across Filtering Development*

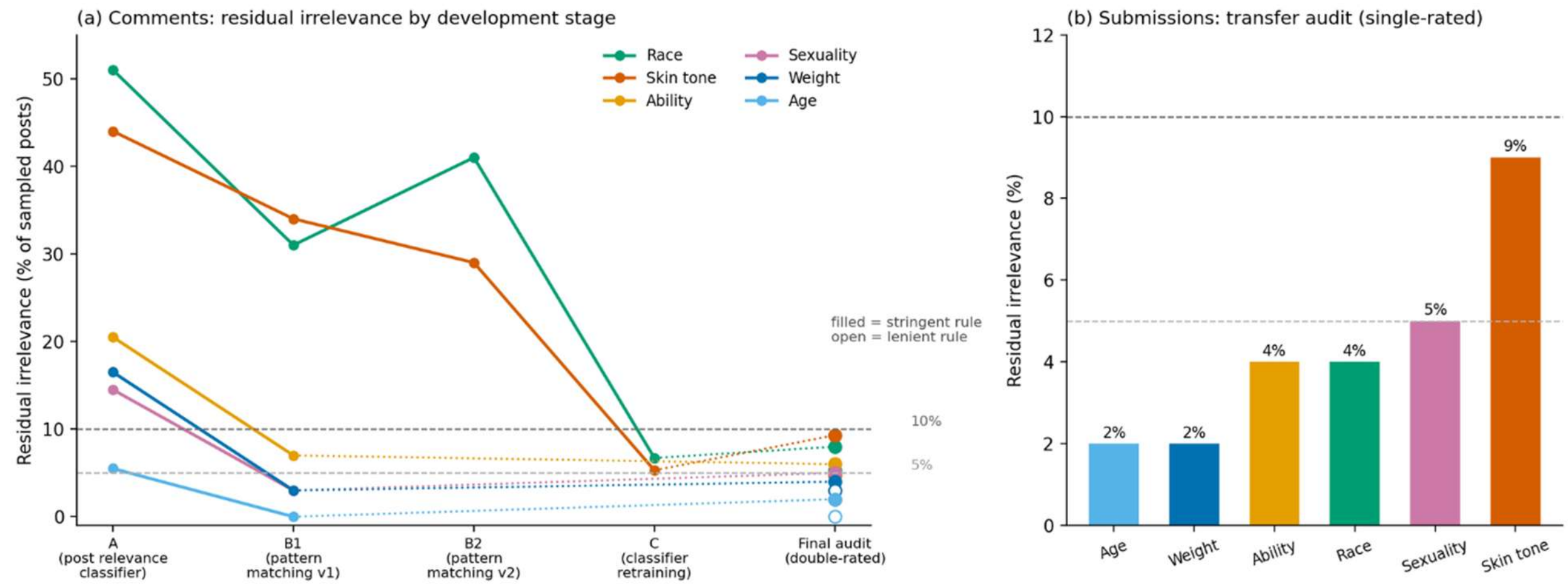

*Note.* (a) Comments: single-rater development samples at Stages A (post relevance classifier), B1/B2 (complex pattern matching), and C (classifier retraining), with the final double-rated audit shown under the stringent (filled) and lenient (open) rules; ability's final audit follows an additional pattern pass (Stage D). Dashed lines mark 5% and 10%. (b) Submissions: single-rater transfer audit of the comment-tuned pipeline.

**Table C3**

*Residual Irrelevance Rate by Development Stage*

| Distinction | A: Post-classifier ($k$ = 200) | B1: Patterns v1 ($k$ = 100) | B2: Patterns v2 ($k$ = 100) | C: Post-retraining ($k$ = 150) | D: Submissions patterns ($k$ = 100) |
|---|---|---|---|---|---|
| Ability | 20.5% | 7.0% | — | — | 4.0% |
| Age | 5.5% | 0.0% | — | — | — |

| **Distinction** | **A: Post-classifier (*k* = 200)** | **B1: Patterns v1 (*k* = 100)** | **B2: Patterns v2 (*k* = 100)** | **C: Post-retraining (*k* = 150)** | **D: Submissions patterns (*k* = 100)** |
|---|---|---|---|---|---|
| Body weight | 16.5% | 3.0% | — | — | — |
| Race | 51.0% | 31.0% | 41.0% | 6.7% | 4.0% |
| Sexuality | 14.5% | 3.0% | — | — | — |
| Skin tone | 44.0% | 34.0% | 29.0% | 5.3% | 9.0% |

*Note.* Rates are based on single-rater development samples on comments. Ability, race, and skin tone subsequently received a final pattern pass to ensure transfer to submissions (Stage D).

## C.4 Corpus-Level Validation Procedures

### *C.4.1. Localized Event Analysis*

The nine state-level marriage-equality ballot measures falling within ISAAC's window were California's Proposition 8 (2008), Arizona's Proposition 102 (2008), Florida's Amendment 2 (2008), Maine's Question 1 (2009 and 2012 iterations), North Carolina's Amendment 1 (2012), Maryland's Question 6 (2012), Washington's Referendum 74 (2012), and Minnesota's Amendment 1 (2012). For each election, sexuality-relevant posting volumes during the election month and the month following were measured against that state's own historical trailing baseline window (months −14 to −3). To control for widespread national interest and account for platform-wide traffic surges, these voting-state ratios were benchmarked against an identical ratio calculated simultaneously for all non-voting jurisdictions. The baseline excludes the two months before each vote, when campaign activity already elevates discussion, and spans a full year so that each calendar month is represented once, neutralizing seasonality and stabilizing sparse state-month counts. Identical windows were applied to the non-voting placebo states.

### *C.4.2. Sentiment Composite*

To construct the longitudinal validation index used in the comparison against national public opinion ($k$ = 101,888), we systematically sampled 500 comments from every monthly archive of sexuality-related discourse spanning 2007 to 2023. For each sampled entry, the continuous outputs of the three sentiment classifiers (VADER, TextBlob, and Stanza) were standardized into $z$-scores and averaged annually to yield a unified tracking metric.

## C.5 Affective–Moral Profiles Across Distinctions

Figure C2 presents the full affective–moral profile of each social group distinction and the Moralization × Emotion interaction summarized in the main text.

Moral content is rare in unselected communication: estimates from ambient recordings of everyday speech and donated personal social media posts fall between 2% and 5% (Atari et al., 2023). In ISAAC, by contrast, 49% to 74% of the discourse is marked as moralized across group distinctions. The contrast likely reflects the impacts of both data specification and measurement approach: ISAAC's Reddit content is filtered for relevance to social group distinctions that are known subjects of intense normative scrutiny, while Atari and colleagues' (2023) estimates relate to moralization across all posts (including personal updates) in a sample of users that felt comfortable sharing their private social media histories. ISAAC's binary operationalization is also more inclusive than the moral foundation-specific coding employed in the earlier estimation.

**Figure C2**

*Affective–Moral Structure Across Social Group Distinctions*

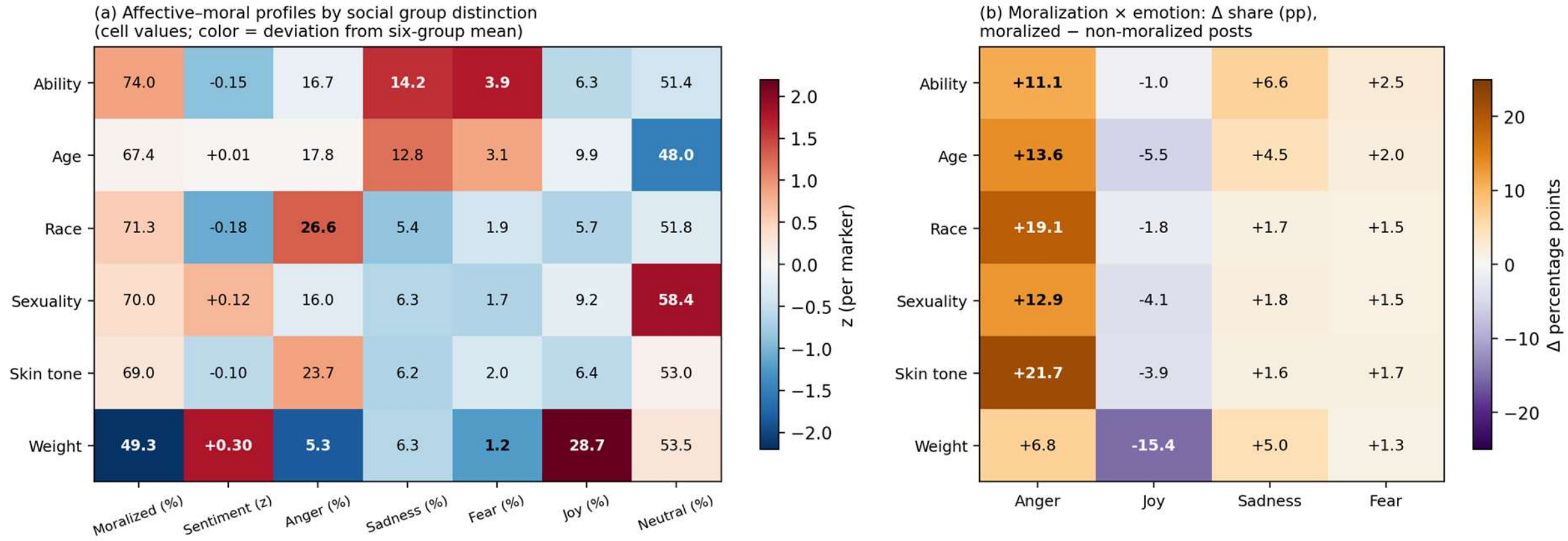

*Note.* Samples are fresh volume-weighted 100,000-post draws per distinction from the final corpus; emotion shares are the mean of the three models' top-label indicators. (a) Profiles: cell values (percentages; sentiment in ensemble z-units), colored by deviation from the six-group mean within each marker. (b) Moralization × Emotion interaction: change in each emotion's share among moralized relative to non-moralized posts, in percentage points.

# Appendix D

# Location Labeling

## D.1. Overview

Reddit has no native geography, and only a small fraction of posts contain explicit location mentions. Constructing usable and generalizable per-user location labels at corpus scale is therefore a significant methodological challenge. The closest prior standard for inferring Reddit users' locations is Harrigian's (2018) SMGeo framework. Its contribution had two parts: (i) a procedure for using textual self-statements ("I live in X", "I'm from Y") to assemble a training set of users with known locations, validated against a curated gazetteer, and (ii) a joint model trained on per-user word, subreddit, and timestamp features that predicts a user's city of residence. SMGeo's labeling scheme is finer-grained than ISAAC's, city-level rather than state-level, and remains the most thoroughly documented attempt to give a large Reddit corpus a geographic dimension.

ISAAC and SMGeo confront the same broad set of constraints; where they differ is in how heavily each constraint weighs against the downstream use case. First, scale: training and inference must scale to the full 2007–2023 Reddit archive (2.29 TB; containing features from millions of annotatable users featured in ISAAC). The original SMGeo feature generation and inference pipeline, aside from relying on a discontinued API, does not extend tractably to more recent years or ISAAC's overall scale. Second, Reddit's geography is heavily skewed toward a handful of large states and toward English-speaking countries. Without explicit corrections, a model that minimizes overall loss collapses onto dominant or rare labels and loses the state-by-state variation that ISAAC's geographic research questions depend on. Where SMGeo trained a flat multinomial classifier and let its training-data filter determine what balance emerged, ISAAC adds explicit class re-weighting and a hierarchical decomposition of the prediction to prevent any class from collapsing. Third, for those same questions, a confidently mistaken label is more costly than an honest abstention; the model should be able to provide tiered responses or outright refuse to predict, unlike the SMGeo alternative. Fourth, ISAAC is intended as an extensible platform given access to newer data, so re-training and re-labeling must be cheap enough that a research group of ordinary resources can do it. This constraint did not bind SMGeo, which was developed as a one-time research output rather than a refreshable corpus. ISAAC's modular LR mixture (see Appendix D.3.1) keeps the platform extension feasible where a joint neural retrain on subsequent corpora would not be.

The discussed constraints lead us to retain Harrigian's labeling-by-self-statement approach (extended to 2023) and keep his three feature views (per-user words, subreddits, and hours of activity), while departing from SMGeo in four respects: a coarser label set (U.S. state or world region, not city); a per-feature logistic-regression mixture in place of a slower single joint model; confidence thresholds with an explicit "unknown" fallback; and an inference resource (label_location) with persistent caching that makes corpus-scale labeling feasible without redundant scans of raw Reddit data.

## D.2. User Labels and Feature Sets

### *D.2.1 User Location Annotation*

The training set was constructed in two stages: candidate sampling of texts where users likely stated where they live, followed by gazetteer-based location resolution. Both stages adapt Harrigian's (2018) procedure to the ISAAC time window (2007–2023).

**Candidate Sampling.** We scanned the full Reddit archive for posts containing self-locating phrases like responses to "where are you from", "where do you live", "where are you living", as well as posts from the subreddit AmateurRoomPorn, where users routinely describe where they live while showing related photographs. We retained only top-level comments and original submissions; replies to other users' self-locating prompts are far more likely to describe someone else's location than the user's own. Only unique authors and posts were retained, and posts containing terms that refer to travel or migration ("move", "moving", "born", "raised", "travel", "trip", "leave", "leaving") were removed. Because our procedures closely match Harrigian's, the 2007–2018 candidate set largely overlaps with the corresponding portion of his sample. The 2019–2023 candidate set is a novel addition.

**Location Resolution.** Surviving candidates were passed through a LocationExtractor module adapted from Harrigian's SMGeo implementation. The extractor scans each post's text against a gazetteer derived from GeoNames, restricted to populated places (GeoNames feature class "P") and administrative regions (feature class "A") with a minimum population threshold of 15,000 to suppress noisy small-place matches, and to entries written in English or Latin script. Candidate strings matching the gazetteer were resolved into structured records of "city, state, country, latitude, longitude" using the Nominatim geocoding API. Robustness measures shaped the resolver: (i) when a full string failed to resolve, we backed off to partial matches on its comma-separated components (a user writing "Springfield, IL, USA" would resolve to "IL, USA" or "USA" if the full string failed); and (ii) subreddit-specific region biases broke ambiguous matches (a "Cambridge" mention in a UK-leaning subreddit resolves to Cambridge, England rather than Cambridge, MA). Users with multiple distinct location mentions were assigned the geometric median of their latitude–longitude pairs, on the grounds that the downstream construct of interest is the user's rough home area rather than any specific point. Finally, the country field was normalized to a canonical lowercase form using an alias list (the United States, for example, appears in raw form as “United States”, “USA”, “U.S.”, etc., all of which collapse to a single canonical “us”). Users with unresolvable country fields (0.5% of the set) were excluded from training.

**Validation.** A random sample of 100 labeled users was hand-reviewed. Excluding users whose conversation context made the location unverifiable at the indicated level, U.S.-vs-non-U.S. classification was correct in 88 of 92 cases (95.7%), state classification within the U.S. in 46 of 51 (90.2%), and region classification outside the U.S. in 36 of 40 (90.0%), as referenced in the main paper.

**Final Sample.** The final annotated set contained 119,019 unique users (68,336 U.S., 50,683 non-U.S.; ratio comparable to Harrigian, 2018). The U.S. and non-U.S. geocoded sets were stored separately for more fine-grained model training (see the Model details below). For user-privacy reasons, the training dataset is not distributed publicly; researchers interested in re-use should contact the corresponding author.

### *D.2.2 User Feature Views*

For each labeled user, we built three feature views derived from the user’s entire Reddit posting history (not the subset of posts curated into ISAAC, because location is a property of the user rather than of any topic of discussion). The three views (words, subreddits, and hours of activity) correspond to the three feature classes used by Harrigian (2018) and follow similar formatting conventions; the substantive change in the present pipeline is downstream, in how they are combined into the inference model (see Appendix D.3.1).

Each view captures a distinct kind of geographic signal. Words carry direct regional linguistic information: local place names, dialect-marked terms, references to local institutions, products, and events. This signal is sparse, noisy, and unevenly distributed (active users produce strong signal, lurkers produce little) but can be highly informative when present. Subreddits capture community affiliation: participation in r/Texas, r/SeattleWA, or r/CasualUK is a far less ambiguous signal than scattered words, but a thinner one (most users participate in relatively few distinct subreddits over their lifetime). Hours of activity, computed as a 24-bin histogram of a user's post timestamps in UTC, serve as a coarse proxy for the user's time zone. The hour signal is weak per user but dense (every post contributes), and it recovers a usable geographic prior for relatively few posts.

The three views therefore differ along three axes that matter downstream: sparsity (how often the view applies per user), noise (how reliably its non-zero entries reflect geography), and discriminative power (how strongly its strongest entries point at one location rather than several). Combining them at the probability level, as we do in Appendix D.3.1, enables each view to contribute optimal information while mitigating performance degradation caused by the others. Of the three views, word features required more detailed preparation, as discussed below.

#### D.2.2.1. Word Features.

User-level vocabularies were built after rigorous preprocessing of user posts to minimize noise and maximize location signal: texts were lowercased and tokenized to alphabetical elements of length at least two. Stopword filtering included NLTK's standard set as well as conversational fillers, generic verbs, social media terms and platform-specific noise (see the complete list below). The remaining tokens were then lemmatized using WordNet to aggregate their contributions across their different forms. Each user's entry in the final vocabulary consisted of a word and its corresponding frequency across all processed submissions and comments from that user.

***D.2.2.1.1. Conversational Fillers***

yeah, yep, ok, okay, hmm, uh, oh, ah, lol, haha, hey, hi, hello, bye, thanks, thank, welcome, please, sorry

***D.2.2.1.2. Generic Verbs and Auxiliaries***

get, got, make, made, do, did, does, done, say, said, go, went, see, saw, look, looking, know, think, feel, want, need, take, put, come, give, use, like, try, let, keep, call

***D.2.2.1.3. Common Adjectives and Adverbs***

good, bad, nice, great, really, very, so, much, many, more, most, some, any, other, same, better, best, new, old, big, small, sure

***D.2.2.1.4. Platform and Social Media Noise***

post, posted, like, share, link, comment, reply, thread, tag, follow, followers, account, login, page, bio, username, profile, hashtag, photo, video, click, url, www, http, https, com, org, net, app, website

***D.2.2.1.5. Common Connectors and Pronouns***

im, ive, dont, cant, isnt, wasnt, youre, theyre, whats, theres, thats, id, wed, hed, shell, hell, weve, couldnt, wouldnt, shouldnt

***D.2.3 Engineering for Scale***

The 2007–2023 Reddit raw dump contains 2.29 TB of compressed data. Constructing per-user feature counts for the labeled training set and, later, for the millions of unlabeled users whose locations the model would predict, required a streaming pipeline that never holds more than a tractable slice of the data in memory at one time. We addressed the high number of users with deterministic hash bucketing: each user is assigned to one of a fixed number of buckets using an MD5 hash of their username, and per-bucket partial counts are flushed to disk periodically as the pipeline streams through posts. Because the hash is deterministic, all a user's partial counts land in the same bucket, and a final merge pass per bucket combines them into a single per-user record without any cross-bucket coordination. The same approach was taken for all feature views. Robust intermediate products ensured that incomplete runs were not wasted, and parallelism was introduced to allow for data processing in a matter of days on a single workstation. This is where ISAAC's approach diverges most sharply from Harrigian's. SMGeo's reference implementation assumes a corpus that can be held in memory and a joint model that can be trained in a single pass. Our adjusted approach uses the same families of features that Harrigian validated, on a corpus two orders of magnitude larger, with workstation-grade resources.

## D.3. Location Inference

### *D.3.1 Model Design*

ISAAC's location model is a weighted mixture of logistic regressions over the three feature views (words, subreddits, hours), applied hierarchically to predict U.S.-vs-non-U.S. for every user, then U.S. state for U.S. users or world region for non-U.S. users. Each of the four design choices below answers one or more of the constraints introduced in Appendix D.1, and each marks a departure from SMGeo.

**Mixture (Rather than Joint) Model.** The feature views differ in sparsity, in scale, and in how cleanly they carry geographic signals. A joint model (Harrigian's choice) absorbs all three into a single representation but, in doing so, lets the densest, noisiest signal, i.e. words, dominate the others. We instead trained separate logistic regressions on word features on one hand, and subreddit and hour features on the other (which we refer to as the "structured" features). We then combined their per-class probabilities at inference time with mixture weights selected through held-out grid search. The weighting was selected so that the final mixture's top-1 accuracy on a masked held-out set, one in which explicit location terms have been blanked out of the text (e.g., "I live in Boston" → "I live in [MASK]"), was maximized. This is a stronger generalization test than ordinary held-out accuracy because it forces the model to earn its predictions from broader linguistic signature rather than from picking up the same self-statements that produced the training labels. The mask-versus-unmask gap is the cleanest measure we have of how much the model relies on surface mentions, and it is reported in the Masked column of Table 4 in the main text.

**Logistic Regression (Rather than a Deep Model).** Logistic regression satisfies all four constraints from Appendix D.1 simultaneously. It scales easily at training time to more than a hundred thousand labeled items, and at inference time to the millions of users in ISAAC. It produces class scores with dependable rank ordering at the higher levels of the hierarchy, which we need for falling back to less granular labels and "unknown" as needed. It is straightforwardly retrainable on extended corpora, satisfying ISAAC's platform extension goal. Class imbalance is addressed by re-weighting the loss inversely proportional to label frequency and prevents the optimizer from collapsing onto majority classes (California, New York, and Texas for the state-level task; the U.S. itself for the top-level task). L2 regularization with strength C = 3.0 was chosen after light grid search; this value leaves enough slack for the model to track infrequent state-marked vocabulary without overfitting to the label distribution's long tail.

**Hierarchical Conditioning.** A flat 52-way U.S. user classifier (the 51 state-level labels plus an unknown class) is doubly disadvantaged: its class imbalance is far more severe than the binary U.S.-vs-non-U.S. task's, and its prediction problem is also harder. We instead trained three separate classifiers: one binary U.S.-vs-non-U.S., one 51-class state classifier conditional on U.S., and an additional 4-class region classifier conditional on non-U.S. We applied them in cascade. Held-out performance shows that the cascade earned its keep: U.S.-vs-non-U.S. was robust both with and without explicit location mentions in the text ($F_1$ above 0.90 in both conditions); region was similarly robust ($F_1$ around 0.80); only the state-level task showed substantial degradation under masking, which is what motivates the confidence-aware fallback below (see Table 4 in the main text).

**Confidence Thresholds and Abstention.** For ISAAC's geographic research questions, abstention is more useful than a confident error. The labeled output for each user was then determined by three threshold rules applied to the mixture's class probabilities. First, the top-level (U.S.-vs-non-U.S.) prediction must exceed a probability of 0.60 to be accepted; otherwise, the user receives the label UNK for "unknown" location. Second, the conditional state prediction following a U.S. determination must exceed its nearest competitor by a margin of at least 0.05; otherwise, the model falls back to the top-level rather than guessing a state. Third, the regional prediction after a non-U.S. determination must exceed its competitor by at least 0.10.

The numerical thresholds correspond to comparable evidence regimes given the branching of each task. For the binary top-level decision, the 0.60 cutoff requires the model to show at least a clear 60-40 preference. The regional decision is over four classes, where chance would place each class at 25% and score gaps between candidates are also large; a 0.10 margin between the top region and its runner-up demands a 40% premium, the smallest gap that excludes near-ties in this regime. The state-level decision is over 51 classes, where chance would place each class at 2% and gaps between candidates are correspondingly small; here the 0.05 margin already requires the top class to exceed its closest competitor by 2.5 times the chance level. The thresholds therefore ask for roughly the similar evidence above chance from each subtask, scaled to the difficulty of the prediction it makes[1]. The main text describes the convergent validation of the resulting labels through state-level event-locked analyses (see Results, Localized Event Spikes).

***D.3.2 Training***

Training proceeds in the three steps described in the repository README (train_location_preprocess → train_location_training → train_location_weighting). The hyperparameter choices below, adjustable through the scripts for future adaptations, follow directly from the stated constraints and design choices.

**Preprocessing (train_location_preprocess).** This step converts the per-user JSON Lines feature files into sklearn-ready sparse matrices. The feature representations differ across the three views because each view's signal is shaped differently.

*Words* are converted to TF–IDF features with sublinear term-frequency scaling, smoothed IDF, and L2 row normalization. TF–IDF is the right choice for words because it down-weights tokens that appear across many users (common words carry little regional information) and up-weights tokens whose distribution is concentrated in a few users (the candidate regionalisms). Sublinear TF compresses the influence of a single user who happens to use a regionalism many times. L2 normalization rescales each user's row to unit length, which strips out the effect of how *much* the user posted and leaves only the *distribution* of what they posted, as total post volume is not itself a geographic signal. The vocabulary is capped at the top 50,000 tokens by document frequency from a candidate pool of 250,000 (with a minimum DF ≥ 2), to suppress noise tokens that appear in a few users.

---

[1] These margins are expressed on the model's raw score scale, which is compressed relative to true probabilities. See sections 3.6–3.7 of this appendix.

*Subreddits* are converted to log1p-transformed frequencies with L1 row normalization. The log1p transform compresses the heavy-tailed distribution of subreddit counts so that a user who posted 5,000 times in one subreddit does not drown out their participation in other subreddits. L1 normalization (so that each row sums to one) again strips out volume effects, since the geographic signal is in the *mix* of subreddits a user participates in, not in absolute counts. The subreddit vocabulary is capped at the top 20,000 by document frequency (min DF ≥ 2). L2 would be the wrong row normalization here because it would inflate the influence of the single dominant subreddit; L1 expresses the row as a probability distribution over subreddits, which is the form the LR classifier interprets most cleanly.

*Hours* are converted to L1-normalized 24-bin histograms, with each row summing to one. No log1p transform is needed because hour bins are inherently bounded (24 of them) and rarely span orders-of-magnitude differences. L1 normalization is again the better choice because what matters geographically is the *shape* of the daily activity curve rather than its amplitude.

The preprocessing pass also generates a parallel masked words matrix in which explicit location terms have been blanked out; this is what the Appendix D.3.1 mask-validation step uses.

**Training (train_location_training).** For each (task × feature set) combination, three tasks (top, state, region) and two feature sets (words, structural), an sklearn LogisticRegression is fit with the default lbfgs solver, regularization strength C = 3.0, class_weight = “balanced”, and up to 200 iterations. The script logs per-class metrics on both regular and masked held-out sets so that the masking-induced degradation is visible for each subtask.

**Weighting (train_location_weighting).** This step searches a grid of mixture weights, the relative contribution of the structural model versus the words model, sweeping from 0.00 to 0.50 in steps of 0.05, and selects the weight that maximizes top-1 accuracy on the *masked* validation set. The masked validation criterion is deliberate: it is the version of accuracy that best reflects how the model will perform on users whose Reddit history does not include a self-statement. The script reports both regular and masked validation metrics so that researchers can inspect the gap. Because this criterion depends only on which candidate ranks first, it constrains the ordering of the mixture's scores but not their scale; a sweep of the mixture weight confirms that the accuracy-optimal and calibration-optimal weights differ at every level of the hierarchy, and that no weight setting would yield calibrated scores at the most granular level (see Appendix D.3.6).

### *D.3.3 Inference at Scale (label_location)*

A trained location model must be applied to each of the millions of users who appears in ISAAC over its 17-year window. Applying it naively, with one pass through the raw Reddit dump per user, would be computationally infeasible. The label_location resource is the part of the pipeline that closes the gap between the trained model and a labeled large corpus. For efficiency, the deployed labeler combines the two feature models' leading candidates rather than their full class distributions. On held-out users this changes top-1 accuracy by at most 0.4 percentage points and mean confidence by .007, so the held-out figures reported in the main text describe the deployed labels rather than an idealized variant of them. The operational details of the label_location resource are recorded in label_location_internals.md in the repository; here we describe the choices that bear on the validity and reproducibility of the labels.

The resource processes one curated monthly ISAAC file at a time. For each user appearing in that file, it gathers a window of the user's broader Reddit activity, applies the trained model, and writes a location label, the model's confidence (location_prob), and the runner-up label (contender_location) as well as its corresponding probability. Five design choices shape how it does this, each tied back to one of the constraints from Appendix D.1.

*User samples are drawn from a window of months around the curated month.* Rather than limiting features to posts within ISAAC or reading every raw file ever produced for every user, the resource reads a controlled window of complete Reddit months centered on the curated month, by default 30 months on either side for years 2007–2019 and 20 months for 2020–2023. Within that window, the script accumulates up to a fixed cap of posts per user (50 for 2007–2019, 25 for 2020–2023). These caps were chosen so that the *distribution of samples per user* at inference time matches the distribution the model was trained on; a user whose inference-time sample is much larger than any training user's would receive predictions outside the range the model was fitted on. The relaxation for 2020–2023 accommodates the much larger monthly archives in recent years while maintaining reliability.

*Two persistent on-disk caches eliminate redundant work.* The first cache stores final per-user labels, so that a user who has already been labeled is not re-labeled when they reappear in a later curated month. The second cache stores per-user feature counts indexed by raw file, so that re-scanning the same user across overlapping windows is unnecessary. Both caches are shared across the six social group distinction labeling runs (sexuality, race, age, ability, weight, skin tone), with the consequence that labeling a second distinction is substantially cheaper than the first due to authorship overlap and labeling the sixth is almost free. However, these caches require significant storage space on the device.

*Concurrency and crash recovery are built in.* Parallel labeling jobs cannot corrupt each other's state because the caches use idempotent write rules: if two jobs label the same user, the one with the higher confidence wins. The output is written in source-row order, with periodic flushes, so that a job killed mid-run can resume from the last completed row rather than starting over. Running many jobs in parallel and running one serially produce identical labels.

*A deduplication step prevents double-counting.* The curated ISAAC file is itself a subset of the raw Reddit dump for the same month, so without deduplication a post that appears in both the curated file and the broader monthly window would contribute to the user's feature counts twice. A simple post-ID check eliminates the double count.

*A minimum-sample-size rule protects against thin users.* Users whose Reddit history within the window yields fewer than 10 usable posts are labeled UNK directly rather than being scored by the model. Their predictions would rest on far less evidence than any training user's, and abstention is more useful than a poorly supported guess.

The peak memory footprint for a single month's labeling is up to 25 GB. A full ISAAC labeling run is suggested only on a CPU-heavy cluster with good memory resources.

***D.3.4 Label Distribution***

Figure D1 shows the location label distributions that cleared the conservative decision threshold to be assigned as user labels in the corpus, with hierarchy levels in different panels.

ISAAC accounts' relative U.S. state share (1,724,720 accounts) aligned with Census Bureau's Vintage 2024 population estimates (DC included): Pearson's $r = .93$ ($r = .88$ if weighted by post frequency). No major differences across distinctions were observed, range from $r = .91$ to $r = .93$. The median state sat at 1.05 times its population share, but only 23 of 51 fell within 0.8 to 1.25×. The residual was systematic: over-represented units were small, distinctive states (e.g., WY 4.6×, AK 4.2×, DE 3.9×, HI 3.2×, UT 3.2×, OR 2.9×); under-represented ones included populous states that share a metro area with a neighbor (NJ 0.48×, NV 0.54×, KS 0.56×).

**Figure D1**

*Estimated Home Location of ISAAC Users*

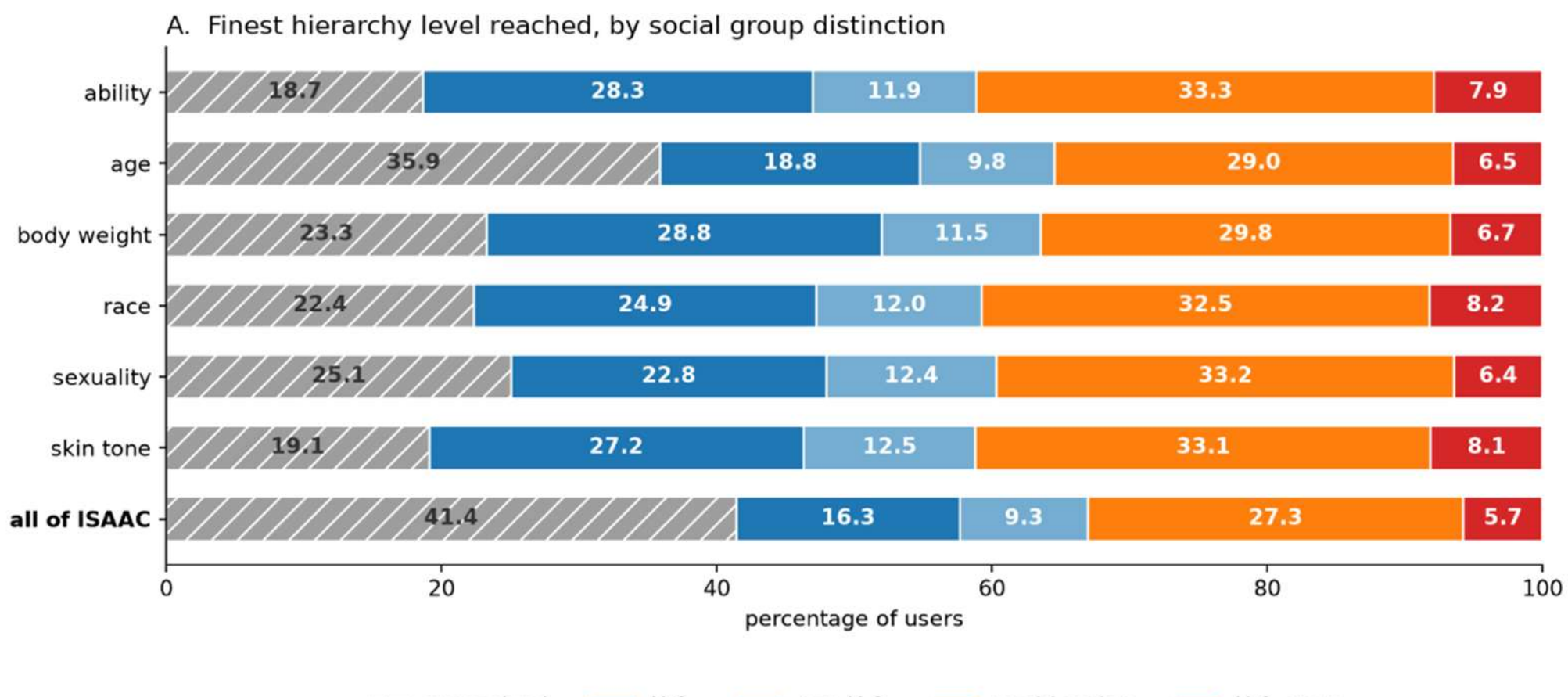

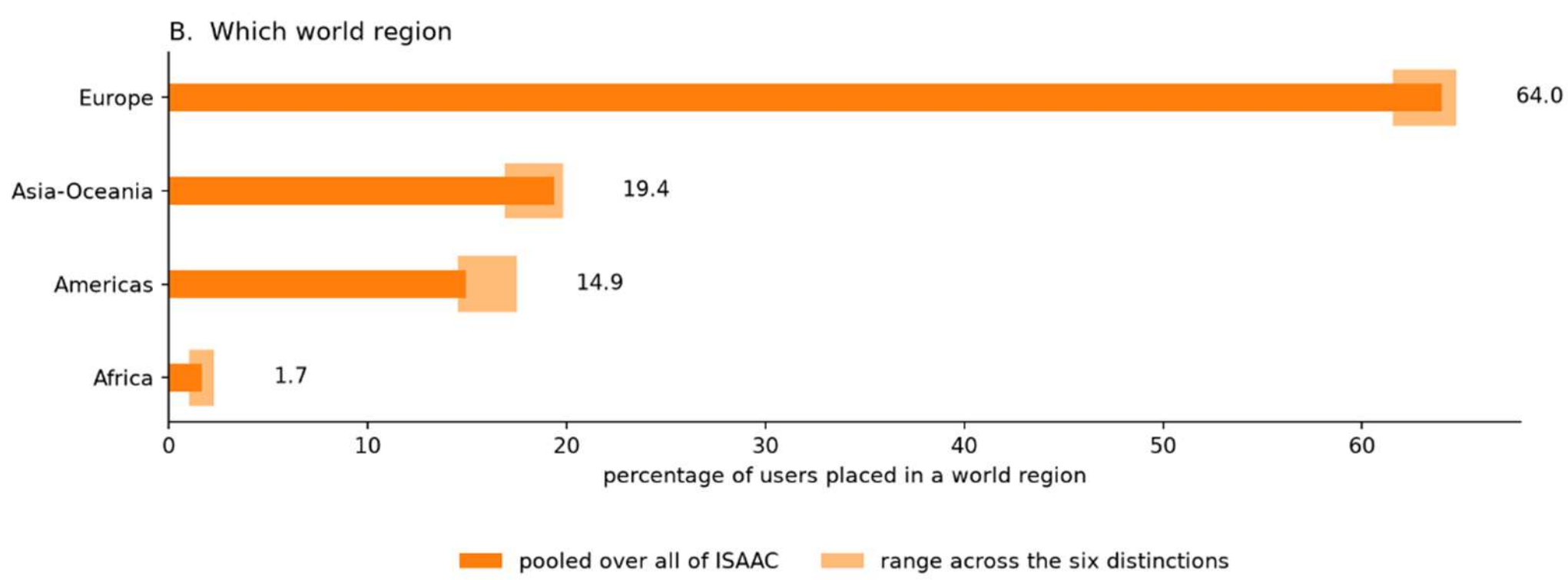

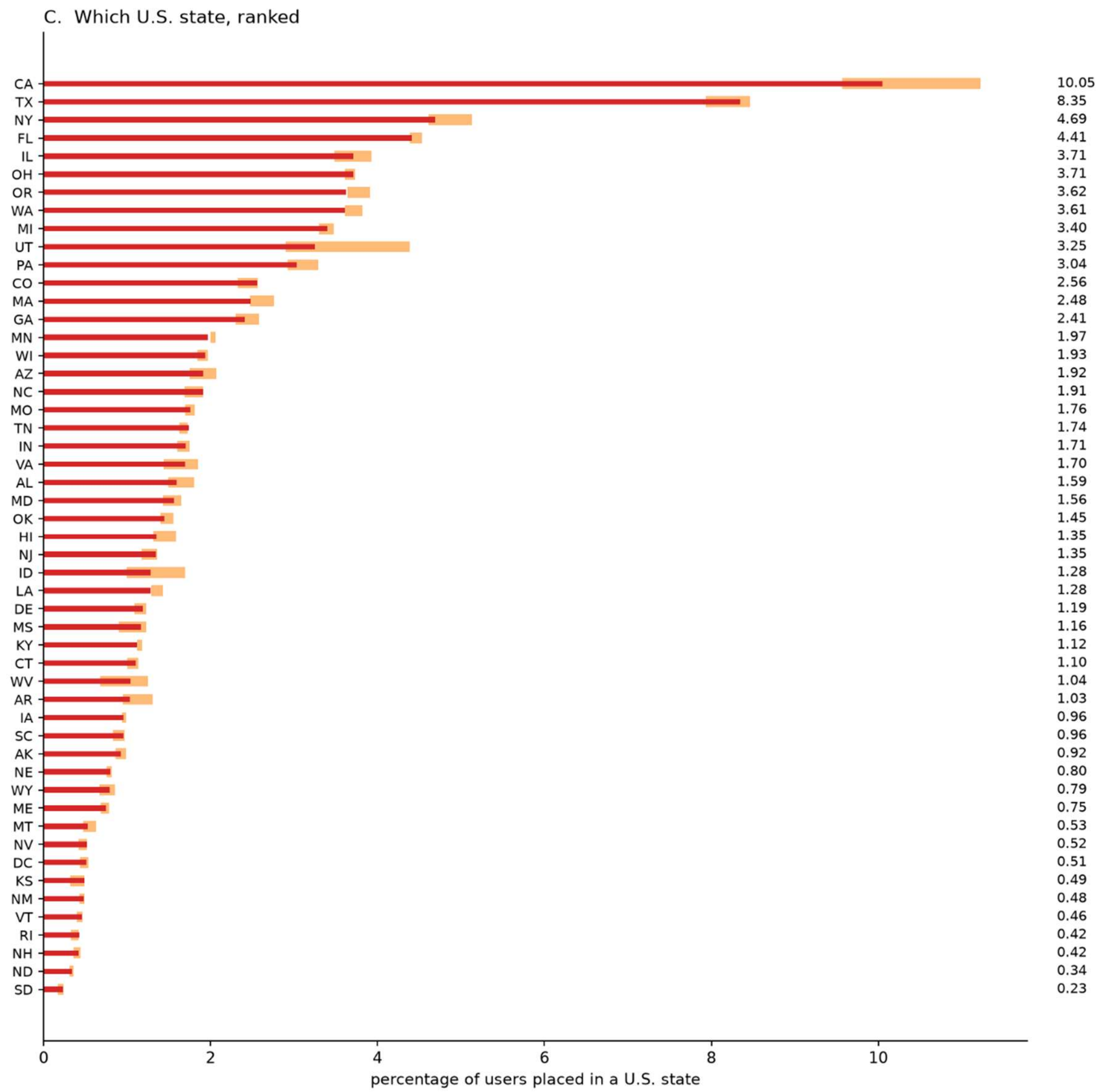

*Note.* (A) Share of accounts resolved to each location hierarchy level, by social group distinction and pooled across the corpus. (B) Distribution of world regions among accounts placed at region level. (C) Distribution of US states among accounts placed at state level, ranked. Solid bars in B and C are pooled over all of ISAAC; shaded bands show the range across the six social group distinctions.

### *D.3.5 Performance*

Held-out performance for the three classifiers under standard and masked conditions is reported in Table 4 of the main text. State-level values under both conditions are shown in Figure D1.

**Figure D2**

*State Model's Evaluation on Held-Out Users*

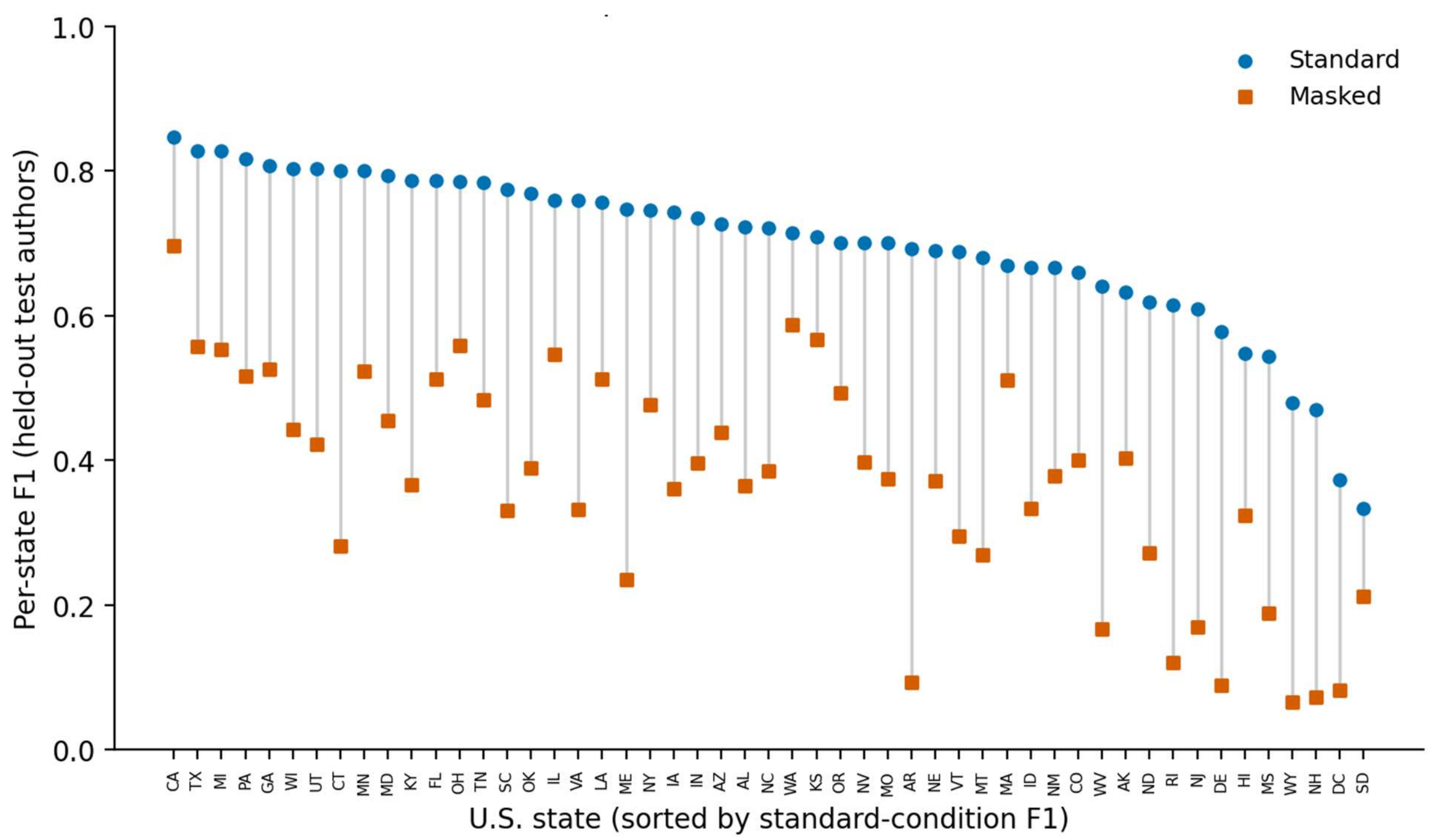

*Note.* Sorted by standard-condition $F_1$.

### *D.3.6 Calibration of the Reported Confidence Values*

To determine whether the model's internal confidence scores, shipped with the corpus, reflect the location labels' true probability of correctness, we evaluated the model's expected calibration error (ECE) across each geographic tier, both for all held-out users at each level and for the thresholded labels the pipeline actually assigns (see Figure D3). Across all geographic tiers, the confidence values remained uniformly conservative. Among all held-out users, the ECE was

.09 for the U.S. versus non-U.S. tier, .16 for the world region tier, and .57 for the state tier; over the emitted labels, the corresponding values were .09, .11, and .52.

For the coarser U.S. versus non-U.S. and world region tiers, the internal confidence score serves as a highly dependable ranking signal. Across the range of confidence scores, empirical accuracy climbed from .78 to .97 for the U.S. versus non-U.S. classification, and from .53 to .92 for world regions, demonstrating strong rank correlations between binned confidence scores and true accuracy ($r = .98$ and $r = .92$, respectively). By contrast, confidence scores were not reliably predictive at the U.S. state level, where accuracy tightly clustered between .73 and .83 across all confidence values ($r = .66$). A key reason is that the decision thresholds during inference have already filtered out most ambiguous cases. A calibration sweep of mixture weights from 0 to 1 confirms that no setting produces calibrated scores at this level; the nominal calibration optimum is the degenerate one that discards the word model entirely and drops accuracy from .75 to .25.

***D.3.7 Confidence Recalibration***

Because the distortion is monotone at the U.S./non-U.S. and region levels, it can be corrected without retraining. We fit an isotonic regression per tier mapping the reported score onto the observed probability that the label is correct, estimated on the validation split of held-out users and evaluated on the test split. Applied to the labels the pipeline emits, this model reduced expected calibration error from .09 to .011 at the U.S./non-U.S. tier and from .11 to .007 at the region tier. Isotonic regression is monotone, so recalibration never reorders users or changes any label; it changes only the interpretation of the score.

The appropriate mapping depends on the user. Fitted under standard conditions it describes users whose text contains explicit location mentions; fitted under masking it describes users who never state where they live. We release both mappings in the primary corpus repository and recommend the masked mapping by default, since most Reddit users do not disclose their location. The mappings are released as location_calibration_maps.json, a table of step-function knots applied by linear interpolation. Calibration holds in aggregate on the population fitted on, namely users whose location was recoverable from explicit self-disclosure. Per-subgroup calibration was not assessed.

No such mapping is released for the state tier given the dissociation from prediction quality. Fitting an isotonic map to that tier collapses it to approximately the tier's base rate: a state label is correct about 77% of the time for users whose text contains explicit location mentions, and about 52% of the time for those it does not. Downstream state-level analyses should therefore treat each state assignment as a categorical label with a known, uniform error rate, rather than weighting individual observations by their reported confidence.

**Figure D3**

*Reliability of the Reported Location Label Confidence Values*

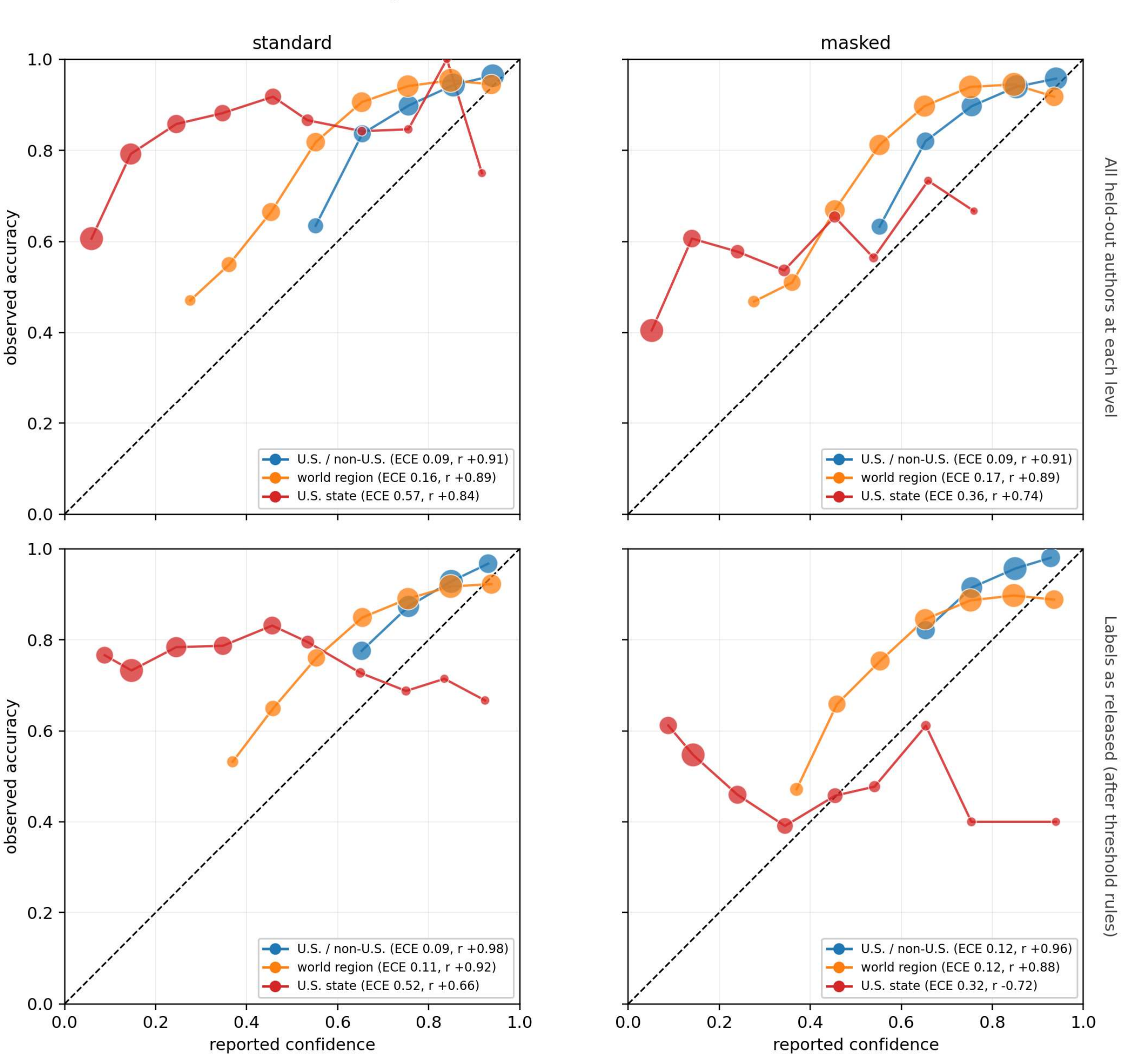

*Note.* Values are shown at the three levels of the hierarchy; the dashed diagonal marks perfect calibration, and curves above it indicate under-confidence. Curves are shown for the standard and masked conditions. The top row scores every held-out user belonging to a level ($n$ = 11,875 / 5,102 / 6,413, as in Table 4); the bottom row scores only the labels the pipeline emits after its threshold rules, which is what the released location_prob column contains. Marker area is proportional to the number of users in the bin, and bins with fewer than five users are omitted.

# Appendix E

## Reusing and Adapting the ISAAC Pipeline

The Method section describes the steps that were completed to construct the released corpus. This appendix documents the same pipeline as a set of reusable resources for readers who wish to recreate ISAAC, extend it, or adapt it to new social groups, constructs, languages, platforms, or time periods. It assumes basic command-line familiarity; full implementation details are provided in the project repository at https://github.com/BabakHemmatian/Illinois_Social_Attitudes, and inquiries may be directed to the first author.

### E.1 Resources and the Command-Line Interface

Each pipeline stage is implemented as a stand-alone resource invocable through a unified command-line interface (cli.py) with a small set of core arguments: the resource name, the target social group, the year range, and the Reddit post type. For example, moralization labels for the sexuality social group can be generated with:

```
python code/cli.py -t comments -r label_moralization -g sexuality -y 2007-2023
```

Table E1 maps each step named in the main text to its resource script. Among the resources, filter_keywords must be run first as the only step interacting with raw data. For other resources, if the order is not customized, they run in the canonical order illustrated in the table. Default development hyperparameters are recorded inside the associated resource scripts.

**Table E1**

*Correspondence Between Main-Text Steps and Pipeline Resources*

| Canonical Order Step | Main text step | Resource script | Purpose |
|---|---|---|---|
| 1 | Keyword matching | `filter_keywords` | Scans the raw Reddit archive for keyword-list matches; the only resource that reads the raw data |
| 2 | Language filtering | `filter_language` | Retains posts identified as English (or another configured language) |
| 3 | Relevance classification | `filter_relevance` | Applies the per-group fine-tuned relevance classifier |
| 4 | Complex pattern matching | `filter_keywords_adv` | Applies per-group regular-expression pattern sets to prune structural false positives |
| 5 | Moralization label | `label_moralization` | Applies binary moralization classifier |
| 6 | Sentiment labels | `label_sentiment` | Applies Stanza, VADER, and TextBlob classifiers |
| 7 | Emotion labels | `label_emotion` | Applies three transformer emotion classifiers |
| 8 | Generalization labels | `label_generalization` | Performs clause segmentation and generates per-clause feature labels |

| 9 | Location estimation | `label_location` | Performs hierarchical location model inference over each user's history |
|---|---|---|---|
| N/A | Merging post types | `organize_types` | Merges separately processed comments and submissions into one time-ordered file per social group |
| N/A | Anonymization | `organize_anonymize` | Replaces usernames with persistent random identifiers. If run as a parallelized batch job, the resource first assigns identifiers to all users in the requested months in a single-process pass, then anonymizes months in parallel against the resulting read-only mapping. |
| N/A | Stratified sampling | `filter_sample` | Draws stratified random samples (by year and per-post keyword-match count) from any stage's output |
| N/A | Interrater agreement | `metrics_interrater` | Computes raw agreement, Cohen's κ, and related statistics on rated samples |
| N/A | Relevance training | `train_relevance` | Fine-tunes a new relevance classifier from rated samples |
| N/A | Location training | `train_location_preprocess / _training / _weighting` | Generates training labels from location self-disclosures; trains the classifiers; fits the feature-class weighting |
| N/A | Verify integrity | `verify_integrity` | Performs stage-aware verification of curated files with structural integrity checks, and anonymization and type-integration reviews. Assumes canonical folders and resource order. Skips checks that are missing inputs and returns "failed" if any files require attention. |

**E.2 Support for Corpus-Scale Computation**

To support development at ISAAC's scale, every resource implements automatic resumption of incomplete runs from a robust per-row provenance column, supports parallelization across processes within a workstation, and supports batch parallelization on high-performance computing clusters via SLURM. The compute-intensive labelers additionally support GPU acceleration. These features apply unchanged to adapted pipelines. A further resource, verify_integrity, can be run after any stage: it re-reads the outputs byte by byte and reconciles row counts and anonymized identifiers against the stage's inputs, catching truncated or corrupted files.

### E.3 Algorithmic Details Behind the Main-Text Descriptions

**Keyword Matching.** The “fast string-matching algorithm” of the main text is the Aho–Corasick algorithm (Aho & Corasick, 1975), via the pyahocorasick implementation, which matches dozens of keywords against arbitrarily long inputs in time linear in the input length, the property that makes exact matching tractable against trillions of raw tokens.

**Language Filtering.** Language identification uses fastText’s pre-trained model (Joulin et al., 2017), chosen for combining near-state-of-the-art identification accuracy with the throughput required for billions of documents.

**Complex Pattern Matching.** The “complex patterns” of the main text are long regular expressions. They are applied with Intel Hyperscan (Wang et al., 2019), a CPU-parallelized multi-pattern regular expression engine, which we further parallelized across workers within a workstation. The per-group pattern sets are included in the repository.

**Relevance Classifier Retraining.** The retrained race and skin-tone classifiers described in the main text as “tuned to be conservative” were fit with an asymmetric penalty in the training loss (misclassifying an irrelevant post as relevant incurs extra cost) and deployed with a decision threshold requiring the model’s estimated probability of relevance to exceed 0.6 (the other four groups use the standard most-likely-class rule). Both choices favor precision over recall, because the operational target was the residual irrelevance rate observed in stratified samples of the re-leased corpus rather than balanced performance.

### E.4 Adaptation Recipes

**A Different Data Source.** Edit the raw-data reader and column expectations in filter_keywords. All downstream resources consume the same intermediate CSV schema regardless of the original source, so no other change is required.

**A Different Language.** Change the language label retained at the filter_language step.

**Different Social Groups or Constructs.** Supply new keyword files following the repository's naming convention; draw double-rated training samples with filter_sample; check their reliability with metrics_interrater; and fit a classifier with train_relevance, which implements the same training procedure used for ISAAC's six released classifiers.

**A New Location Model.** The three train_location_ resources reproduce the full procedure on formatted, user-supplied data: (a) train_location_preprocess produces the per-user feature set matrices needed for training, (b) train_location_training fits the hierarchical classifiers, and (c) train_location_weighting fits the weighting across words and structured (subreddits + hours) feature classes, producing the held-out evaluations reported for ISAAC. ISAAC's own location training data are not released (see Appendix A); the trained models are provided.

**Extending Past 2023.** The pipeline runs unmodified on any archive of later months that follows the Pushshift schema; extension is therefore conditional only on researcher access to newer platform data (see Data Source on the discontinued distribution of the original archives).

### E.5 Released Assets

Public releases accompanying the corpus comprise (a) the six relevance classifiers (the retrained versions for race and skin tone, initial versions for the other four), together with their rated training samples; (b) the moralization classifier; (c) the generalization suite (clause segmentation and feature classifiers); (d) the location-model checkpoints (without their training data), available by email under the ISAAC Model Use Agreement, and the confidence recalibration mappings, which are distributed with the corpus since they require no model access.; (e) all keyword lists and complex-pattern sets; the default configurations of every resource; and (f) the coding-free demo applications described under Adapting and Extending ISAAC in the main text.

These assets are released under three distinct sets of terms. The pipeline code, keyword lists, pattern sets, and default configurations are released under the MIT License. The relevance, moralization, and generalization models are released under a Creative Commons Attribution 4.0 International License without access restrictions. The location model can be obtained by emailing isaac.corpus.support@gmail.com and accepting the ISAAC Model Use Agreement, which requires acceptance before access, prohibits attempts to re-identify individuals, bars surveillance, profiling, and other harmful applications, and requires citation. The rated training samples distributed with the pipeline are derived from Reddit content and remain governed by the Data Use Agreement (Appendix A).

# Appendix F

## Access Routes for Technical Users

The coding-free website at https://isaac.psychology.illinois.edu serves most research uses. This appendix documents the four additional access routes for users who need scripted, composable, or pipeline-integrated access. All routes expose the same released corpus: one file per social group per month, from January 2007 through December 2023, in Parquet (recommended) and CSV formats, and all are governed by the Data Use Agreement (see Appendix A), with acceptance required before data access on every route.

### F.1 Direct Downloads

The website's direct download endpoint serves the corpus as static files catalogued by a published manifest (manifest.json) listing every file with its size and row count. Researchers can retrieve files for fast download through Globus with any HTTPS tool and use them directly in standard data systems (e.g., DuckDB, Spark) without intermediate conversion. Instructions are provided to registered users on the website's direct-download page.

### F.2 SQL Query Playground

For filtered or aggregated extractions that do not warrant a download, the website includes an SQL query playground that runs entirely in the user's browser: queries execute against the published Parquet files over HTTP through an in-browser database engine (DuckDB-Wasm), fetching only the columns and row segments each query touches; no query or corpus data passes through the project's servers. Files are addressed per social group per month, and a single query can span months and social groups by listing multiple files; the interface provides worked examples and a helper that generates the file list, with a group-identifier column, for a chosen social group and date range. Because computation happens in the browser tab, the playground is suited to aggregation and moderate result sets: results display up to 10,000 rows, with CSV exports up to 100,000, and the interface warns before queries that would scan full post text across many months. Larger extractions are better served by the Python package (F.3) or direct downloads (F.1).

**F.3 Python Package (isaac-data)**

The `isaac-data` package, installable from the standard PyPI repository at https://pypi.org/project/isaac-data/, provides catalog-driven programmatic access in three layers. `catalog()` and `files()` enumerate what exists (categories, months, file sizes, row counts) without transferring corpus data. `load()` reads selected social groups, month ranges, and columns directly into a pandas DataFrame. For Parquet files only the requested columns are transferred over the network, so loading two columns from a multi-hundred-megabyte file moves only a few megabytes. Passing `n=` draws a reproducible stratified sample spread equally across the selected months, uniform within each, matching the website's sampling behavior, and a size guard refuses accidentally very large selections. `download()` fetches whole files to disk with resumable, cached transfers for offline or bulk work. A command-line interface (`isaac-data info / ls / download`) mirrors these functions. The package presents the Data Use Agreement on first data access and records acceptance locally, with best-effort updating of the acceptance records on the server as needed; non-interactive environments can accept ahead of time. It supports Python 3.9 and later on Windows, macOS, and Linux.

**F.4 HuggingFace Datasets and Models Hub Interfaces**

A gated HuggingFace mirror of the corpus offers one configuration per social group and integrates with the `datasets` library for convenient script-based access in Python (https://huggingface.co/datasets/BabakScrapes/isaac-reddit): researchers can stream a social group without a full download or materialize it locally with `load_dataset()`, making ISAAC directly usable in machine-learning and AI pipelines. Requesting access requires accepting the Data Use Agreement and agreeing to cite the project; approval grants token-based access.

ISAAC's in-house models are found in HuggingFace Models Hub with the following URL formats: https://huggingface.co/ISAAC-corpus/isaac-relevance-<social distinction> for relevance classifiers, https://huggingface.co/ISAAC-corpus/isaac-<moralization/generalization/generalization-segmentation> for labelers. Visiting https://huggingface.co/ISAAC-corpus offers an up-to-date, clickable list of the models. Access to the location model weights can be requested by emailing isaac.corpus.support@gmail.com and agreeing to the ISAAC Model Use Agreement.

**F.5 Implementation Notes**

The website's backend is implemented in Python and Shell and its frontend in NodeJS; given the corpus's scale, the backend is optimized for efficient sampling and compression. Both codebases are publicly available under the MIT License for inspection and further development, at https://github.com/BabakHemmatian/ISAAC_Sampler (frontend) and https://github.com/BabakHemmatian/ISAAC_Sampler_Backend (backend). The source of the isaac-data package (F.3) and the tooling used to build and upload the HuggingFace dataset mirror (F.4) are maintained in the pipeline repository at https://github.com/BabakHemmatian/Illinois_Social_Attitudes.